\documentclass[lettersize,journal]{IEEEtran}
\usepackage{amsmath,amsfonts}
\usepackage[linesnumbered,ruled,vlined]{algorithm2e}
\usepackage{array}
\usepackage[caption=false,font=normalsize,labelfont=sf,textfont=sf]{subfig}
\usepackage{textcomp}
\usepackage{stfloats}
\usepackage{url}
\usepackage{verbatim}
\usepackage{graphicx, multirow}
\usepackage{cite}
\usepackage{import}
\usepackage{comment}
\usepackage{color, colortbl}
\usepackage{enumerate}
\usepackage{amsthm}
\usepackage{bm}
\usepackage{amssymb}
\usepackage{booktabs}
\usepackage{siunitx}

\theoremstyle{definition}
\newtheorem{definition}{Definition}
\newtheorem{proposition}{Proposition}
\newtheorem{remark}{Remark}

\newtheorem{theorem}{Theorem}

\newtheorem{lemma}{Lemma}

\definecolor{B}{rgb}{0.81, 0.95, 1}
\definecolor{Y}{rgb}{1, 1, 0.8}
\definecolor{G}{rgb}{0.85, 0.96, 0.8}

\begin{document}

\title{Learning Rhythmic Trajectories with Geometric Constraints for Laser-Based Skincare Procedures}

\author{Anqing Duan, Wanli Liuchen, Jinsong Wu, Raffaello Camoriano,~\IEEEmembership{Member,~IEEE},\\ Lorenzo Rosasco, and David Navarro-Alarcon,~\IEEEmembership{Senior~Member,~IEEE}
\thanks{This work is supported in part by the Research Grants Council (RGC) of Hong Kong under grants 15212721 and 15231023, and in part by the Jiangsu Industrial Technology Research Institute Collaborative Funding Scheme under grant ZG9V. \textit{Corresponding author: David Navarro-Alarcon.}}
\thanks{A. Duan, W. Liuchen, J. Wu, and D. Navarro-Alarcon are with the Department of Mechanical Engineering, The Hong Kong Polytechnic University, Hong Kong. (e-mail: aduan@polyu.edu.hk, chenwanli.liu@connect.polyu.hk, jinsong.wu@connect.polyu.hk, dnavar@polyu.edu.hk).}
\thanks{R. Camoriano is with the Visual and Multimodal Applied Learning Laboratory, Politecnico di Torino, Turin, Italy (e-mail: raffaello.camoriano@polito.it).}
\thanks{L. Rosasco is with DIBRIS, Università degli Studi di Genova, Genoa, Italy, with Laboratory for Computational and Statistical Learning (IIT@MIT), Istituto Italiano di Tecnologia and Massachusetts Institute of Technology, Cambridge, MA, United States, and also with Machine Learning Genoa (MaLGa) Center, Università di Genova, Genoa, Italy (e-mail: lrosasco@mit.edu).}
}

\markboth{}
{Duan \MakeLowercase{\textit{et al.}}: Learning Rhythmic Trajectories from Demonstrations}


\maketitle

\begin{abstract}
The increasing deployment of robots has significantly enhanced the automation levels across a wide and diverse range of industries. 
This paper investigates the automation challenges of laser-based dermatology procedures in the beauty industry; This group of related manipulation tasks involves delivering energy from a cosmetic laser onto the skin with repetitive patterns. 
To automate this procedure, we propose to use a robotic manipulator and endow it with the dexterity of a skilled dermatology practitioner through a learning-from-demonstration framework.
To ensure that the cosmetic laser can properly deliver the energy onto the skin surface of an individual, we develop a novel structured prediction-based imitation learning algorithm with the merit of handling geometric constraints. 
Notably, our proposed algorithm effectively tackles the imitation challenges associated with quasi-periodic motions, a common feature of many laser-based cosmetic tasks.
The conducted real-world experiments illustrate the performance of our robotic beautician in mimicking realistic dermatological procedures; Our new method is shown to not only replicate the rhythmic movements from the provided demonstrations but also to adapt the acquired skills to previously unseen scenarios and subjects.
\end{abstract}


\section{Introduction}\label{introduction}
\begin{figure}
    \centering
    \includegraphics[width=\linewidth]{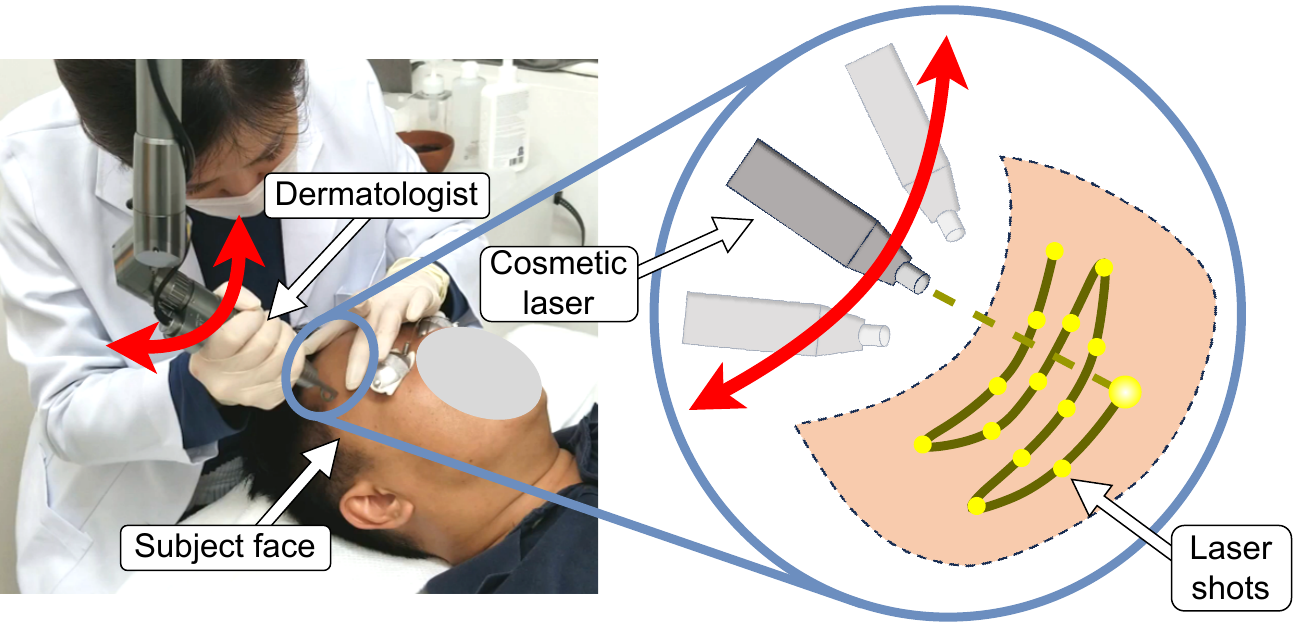}
    \caption{Illustration of a skin photorejuvenation procedure where a dermatology practitioner rhythmically manipulates a cosmetic laser handpiece that fires multiple laser shots onto the subject's skin surface.}
    \label{fig:illurobot}
\end{figure}

Due to its rapidly advancing nature, robotics has drastically changed many aspects of our lives. 
In just a couple of decades, we have witnessed how many labor-intensive industrial processes have been upgraded by the use of robots.
Following this spirit, in this work, we investigate the development of a new generation of robotic systems that automate laser-based cosmetic procedures in the beauty industry \cite{muddassir2021robotics}.
The manipulation tasks under consideration involve the controlled delivery of energy from a cosmetic laser onto the subject's skin with (quasi-)periodic patterns, a procedure that serves multiple purposes, e.g., to improve the aesthetic condition of the skin, decrease the visibility of scars and freckles, remove hair from legs and armpits, to name a few cases \cite{baumann2009cosmetic}.
Fig. \ref{fig:illurobot} depicts the so-called skin photorejuvenation task, where a human practitioner ``rhythmically'' moves the laser handpiece over the area of interest to uniformly deliver thermal energy via pulsating laser shots \cite{9955367}.
This photorejuvenation task (as well as other related laser-based procedures in cosmetic dermatology) can potentially be automated by robots capable of synthesizing skilled (quasi-)periodic motions.
Our goal in this paper is precisely to develop this type of system.

Conventional methods for programming a desired robot behavior may not entirely capture the required skills to perform a dermatology procedure, as these approaches typically hardcode specific motions for the robot, which are difficult to adapt to new situations and subjects. 
To this end, we employ the framework of \textit{learning by demonstration}, also known as robot imitation learning, which offers a convenient paradigm to seamlessly transfer the motion skills from non-expert robot users, such as dermatology practitioners, to a robot by providing the user's motion data~\cite{ravichandar2020recent}.
The entire process of learning by demonstration typically comprises three phases~\cite{10226460}: a \textit{demonstration} phase, where motion data is collected to teach robots the desired behavior, a \textit{reproduction} phase, where robots reproduce the learned skills, and an \textit{adaptation} phase, where robots adapt the learned skills to operate in a different environment from the initial demonstration. 

From an algorithmic perspective, studying representations for motor skills holds significant importance in facilitating robot imitation learning.
The well-established \textit{movement primitives} emerges as a powerful tool~\cite{Billard08chapter}.
Once the demonstration is complete, movement primitives are commonly utilized for motion analysis to extract motion patterns from experts' demonstrations.
Subsequently, during the reproduction and adaptation phases, the learned movement primitives will be employed to accomplish robot tasks.

\emph{Our Contribution.}
Several technical challenges arise when transferring laser manipulation skills from a human expert to a robot.  
The main complications are twofold: (\romannumeral 1) extracting the laser's distinct motion patterns that entail the laser spots rhythmically evolving on a subject's skin surface, and (\romannumeral 2) adapting the extracted motor skills for subsequent treatment of a new subject. 
Accordingly, the original contributions of this paper are outlined as follows:
\begin{itemize}
    \item Development of a novel imitation learning algorithm that can generate rhythmic motion with geometric constraints for laser-based cosmetic procedures.
    \item Derivation of a motor skill adaptation strategy for treating a new subject's face that is absent in the demonstration.
\end{itemize}
Specifically, compared with state-of-the-art methods, our proposed imitation learning algorithm is characterized by:
    \begin{itemize}
      \item Allowing for imitation of trajectories that involve geometric constraints. 
      \item Supporting the acquisition of periodic or quasi-periodic motion patterns.
      \item Eliminating the need for manual configuration or fine-tuning of basis functions.
    \end{itemize}
Our proposed new algorithm can be readily utilized for a variety of imitation learning problems that occur in either the Euclidean space or in the presence of other types of manifold constraints. 
This versatility expands its potential applications beyond the specific context of cosmetic dermatology.

\emph{Organization.}
The rest of the paper is organized as follows:
Sec. \ref{relate} presents the related work; Sec. \ref{imitation} describes the developed learning from demonstration algorithm; Sec. \ref{adaptation} derives the skill adaptation approach; Sec. \ref{experiments} numerically and experimentally validates the proposed methodology; Sec \ref{conclusion} concludes the paper.
Notation is provided in Appendix~\ref{Notation}.

\section{Related Work}\label{relate}
\textbf{Geometric trajectory generation.} 
During photorejuvenation, the robot arm needs to generate laser traces on a subject's face.
It is thus required to ensure that the traces adhere to the geometric constraints imposed by the facial surface.
To achieve geometric trajectory generation, multiple perspectives can be referred to, such as mapping 2D curves onto the 3D surface~\cite{10114055, liu2021robust, jafari2020surface} and sampling~\cite{qureshi2021constrained} or neural networks-based motion planning~\cite{tiboni2023paintnet}.
Compared to the aforementioned approaches, our approach tackles the problem from an imitation perspective. 
It overcomes the distortion issue commonly associated with the mapping strategy. 
Moreover, in comparison to sampling-based methods, our approach exhibits lower computational complexity. 
Particularly, the imitation framework offers a user-friendly interface for non-expert robot users, making it well-suited for dermatologists to transfer their motion skills to a robot arm.

\textbf{Imitation learning on Riemannian manifolds.}
Due to the prevalence of Riemannian structures in robotics, such as orientation, manipulability, and joint stiffness, extensions of imitation learning to Riemannian manifolds have been gaining increasing popularity, see e.g.,~\cite{liu2022robot, abu2020geometry, calinon2020gaussians, rozo2022orientation}. 
In contrast to previous strategies, we propose to address geometric constraints through the use of \textit{structured prediction}, which has proven to be an effective tool for enabling robot movement imitation~\cite{doi:10.1177/02783649231204656}.
Furthermore, in this paper, we emphasize the significant role of kernels, which form the foundation of our employed non-parametric imitation strategy. 
In particular, we highlight the flexibility of our framework in terms of capturing motion patterns by leveraging various types of kernel functions.

\textbf{Geodesic distance calculation.}
Typically, the calculation of geodesic distances constitutes a crucial component in imitation learning on manifolds.
The geodesic distance is commonly available as a by-product of dimensionality reduction methods such as Isometric Mapping (ISOMAP)~\cite{tenenbaum2000global} or Multi-Dimensional Scaling (MDS)~\cite{shamai2018efficient}.
Also, physics-inspired methods, such as the heat diffusion equation~\cite{crane2013geodesics} and the Eikonal equation~\cite{jeong2008fast}, offer an alternative approach to calculating geodesic distances.
In comparison to the aforementioned approaches, we propose to approximate a surface using osculating spherelets, which have the merit of analytic expressions for geodesic distances. 
Consequently, it is efficient to calculate the geodesic distance between two manifold points, without iteratively searching for the shortest path or solving differential equations. 

\textbf{Rhythmic movement modeling.}
Another related field is the generation of rhythmic movement.
Noticeably, parametric imitation is usually achieved by cyclic basis functions~\cite{yang2022learning} while non-parametric methods typically utilize periodic kernels~\cite{huang2020toward}, like our case.
Importantly, we explicitly provide definitions for (quasi-)periodic trajectories under geometric constraints.  
Besides, signal processing techniques, such as the Fourier transform, can also be applied to capture quasi-periodic motion patterns~\cite{li2020learning}. 
However, these approaches disregard the geometry-structured rhythm that our method emphasizes.

\section{Imitation of Geometric Trajectories\\ with Rhythm}\label{imitation}
In this section, we provide the technical details of the developed imitation algorithm. 
First, we present the strategy of motion imitation under geometric constraints by structured prediction (Sec. \ref{sec_structpred}).
Then, we illustrate the determination of the kernel functions for motion pattern capture (Sec. \ref{sec_kernels}),
followed by efficient calculation of loss functions (Sec. \ref{sec_loss}).

\subsection{Motion Imitation by Structured Prediction}\label{sec_structpred}
To begin with, the dataset of the laser spots trajectory demonstrated by a dermatologist is denoted as $\mathbb{T}: \{(t_n, \mathbf{p}_n = [x_n, y_n, z_n]^\intercal) \}_{n=1}^N$, where $t \in \mathcal{T}$ is the time stamp with $\mathcal{T}$ being the input space and $\mathbf{p}$ is the corresponding trajectory position. 
During the treatment, the trajectory of the laser generator cannot freely evolve.
Rather, it is required to be constrained on the inflated surface at a certain height above the subject's face.

\begin{remark}
All the trajectory points from the demonstrated dataset $\mathbb{T}$ lie on the surface manifold $\mathcal{S}$, i.e., $\mathbf{p}_n \in \mathcal{S}$ with $\forall (t_n, \mathbf{p}_n) \in \mathbb{T}$.
\end{remark}

The inherent constraint underlying the demonstrated trajectories poses a unique challenge in applying existing imitation learning algorithms due to the Riemannian metric associated with the surface manifold.  
It is thus imperative to devise geometry-aware movement primitives, which is exactly the objective of this paper.
As a powerful tool to deal with supervised learning, structured prediction is especially competent at handling cases where outputs possess geometrically rich structures~\cite{bakir2007predicting}.
In this paper, we propose to tackle the issue of learning (quasi-)periodic trajectories for skin photorejuvenation by resorting to the structured prediction framework.

A major strategy employed in structured prediction is the surrogate approach~\cite{ciliberto2020general}, whose core notion can be outlined by: 
1) Embedding the outputs into a linear surrogate space, then
2) Solving the learning problem in the surrogate space, and finally
3) Mapping the solution back to the structured space with a decoding rule.
More formally, the surrogate approach to structured prediction is sketched as follows:
\begin{enumerate}
    \item \textit{Encoding.} Design an encoding rule $\mathbf{c}: \mathcal{S} \rightarrow \mathcal{H}$.
    \item \textit{Surrogate Learning.} Solve the surrogate learning problem $\mathbf{g}: \mathcal{T} \rightarrow \mathcal{H}$ which minimizes $\mathcal{L}(\mathbf{c}(\mathbf{p}_n), \mathbf{g}(t_n))$ with the surrogate loss $\mathcal{L}: \mathcal{H}\times\mathcal{H} \rightarrow \mathbb{R}$, given the surrogate dataset $\mathbb{D}_s: \{t_n, \mathbf{c}(\mathbf{p}_n)\}_{n=1}^{N}$. 
    \item \textit{Decoding.} Obtain the structured output $\mathbf{s} = \mathbf{c}^{-1} \circ \mathbf{g}: \mathcal{T} \rightarrow \mathcal{S}$ with a suitable decoding $\mathbf{c}^{-1}: \mathcal{H} \rightarrow \mathcal{S}$.
\end{enumerate}

\begin{figure}[t]
    \centering
    \includegraphics[width=0.99\columnwidth]{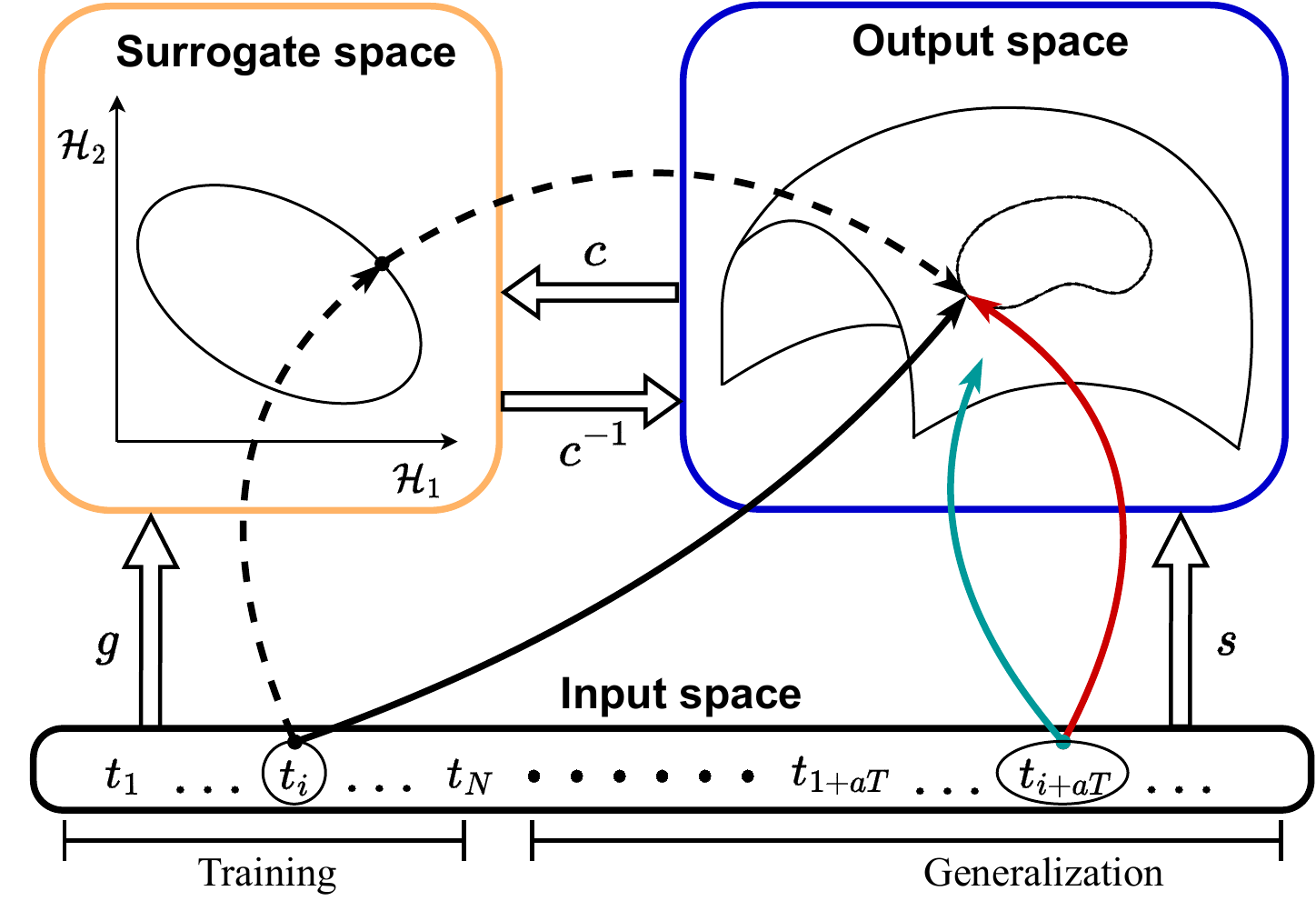}
    \caption{Illustration of making structured prediction for periodic motions using the surrogate approach. Black curves denote the training procedure while red curves and cyan curves denote the generalization procedure using the periodic and squared exponential kernels, respectively.}
    \label{fig:illuKerSP}
\end{figure}

There emerge recent insights on the loss function in the structured space that it implicitly carries a natural corresponding geometric structure.
The exploitation of such a structure allows us to relax the aforementioned surrogate framework further, leading to an implicit formulation.

\begin{definition}[Structure Encoding Loss Function~\cite{ciliberto2020general}]
A loss function $\Delta: \mathcal{S} \times \mathcal{S} \rightarrow \mathbb{R}$ is called a Structure Encoding Loss Function if there exist a separable Hilbert space $\mathcal{H}$ associated with an inner product $\langle \cdot,\,\cdot \rangle_\mathcal{H}$, a continuous feature map $\boldsymbol{\Psi}: \mathcal{S} \rightarrow \mathcal{H}$  and a continuous linear operator $\mathbf{V}: \mathcal{H} \rightarrow \mathcal{H}$ such that for all $\mathbf{p}, \mathbf{p}' \in \mathcal{S}$,
\begin{equation}\label{defSELF}
\Delta(\mathbf{p}, \mathbf{p}') = \langle \boldsymbol{\Psi}(\mathbf{p}) , \mathbf{V}\boldsymbol{\Psi}(\mathbf{p}')\rangle_\mathcal{H}. 
\end{equation}
\end{definition}

By leveraging the ridge regression estimator that minimizes the regularized empirical risk~\cite{alvarez2012kernels}, we can formulate the learning problem in the surrogate space as a regularization problem
\begin{equation}\label{surrogatelearning}
    \mathbf{g} = \underset{\mathbf{g} \in \mathcal{G}}{\operatorname{argmin}} \frac{1}{N} \sum_{n=1}^N \|\boldsymbol{\Psi}(\mathbf{p}_n) - \mathbf{g}(t_n) \|_{\mathcal{H}}^2 + \lambda\|\mathbf{g} \|_{\mathcal{G}}^2,
\end{equation}
where $\mathcal{G}$ is a normed space of functions $\mathcal{T} \rightarrow \mathcal{H}$ and $\lambda >0$ is the Tikhonov hyperparameter.
Considering a reproducing kernel Hilbert space (RKHS) of vector-valued functions and the associated matrix-valued kernel $\bm{\mathcal{K}}: \mathcal{T} \times \mathcal{T} \rightarrow \mathbb{R}^{D\times D}$ where $D$ is the dimension of $\mathcal{H}$, the solution to \eqref{surrogatelearning} is obtained by employing the \textit{representer theorem} within the vector-valued setting~\cite{alvarez2012kernels}
\begin{equation}\label{rkhspre}
\mathbf{g}(t) = \sum_{n=1}^N\bm{\mathcal{K}}(t_n, t)\boldsymbol{\beta}_n.
\end{equation}
By choosing the matrix-valued kernel to be 
$\bm{\mathcal{K}}(t, t') = k(t, t')\mathbf{I}_{\mathcal{H}}$
with $k: \mathcal{T} \times \mathcal{T} \rightarrow \mathbb{R}$ being a reproducing kernel, the concatenation of the coefficient $\boldsymbol{\beta}_n$ gives
\begin{equation}
    \widetilde{\boldsymbol{\beta}} = \begin{bmatrix}\boldsymbol{\beta}_1^\intercal, \ldots, \boldsymbol{\beta}_N^\intercal
    \end{bmatrix}^\intercal =  \big((\mathbf{K}+\lambda N \mathbf{I}_N) \otimes \mathbf{I}_{\mathcal{H}}\big)^{-1}\widetilde{\boldsymbol{\Psi}}
\end{equation}
where the concatenation of the output vectors is given as
$\widetilde{\boldsymbol{\Psi}} = \begin{bmatrix}
\boldsymbol{\Psi}_1^\intercal, \ldots , \boldsymbol{\Psi}_N^\intercal    
\end{bmatrix}^\intercal$ and $\mathbf{K} \in \mathbb{R} ^{N \times N}$ is defined by $\mathbf{K}_{i, j} = k(t_i, t_j)$.  
For any $t \in \mathcal{T}$, we then have
\begin{subequations}
\begin{align}
    \mathbf{g}(t) &= (\mathbf{k}^\intercal \otimes \mathbf{I}_{\mathcal{H}})\big((\mathbf{K}+\lambda N \mathbf{I}_N) \otimes \mathbf{I}_{\mathcal{H}}\big)^{-1}\widetilde{\boldsymbol{\Psi}}\\
    &=\sum_{n=1}^{N}\boldsymbol{\alpha}_n(t)\boldsymbol{\Psi}(\mathbf{p}_n), \label{eq_gt}
\end{align}
\end{subequations}
where $\boldsymbol{\alpha}(t) = [\boldsymbol{\alpha}_1(t), \ldots, \boldsymbol{\alpha}_N(t)]^\intercal = (\mathbf{K}+\lambda N \mathbf{I}_N)^{-1}\mathbf{k}\in \mathbb{R}^N$ 
and $\mathbf{k} \in \mathbb{R}^N$ is constructed by $\mathbf{k}_i = k(t, t_i)$.

Leveraging the structure of the SELF loss function in the expected risk, the following decoding rule can be introduced:
\begin{equation}\label{eq_cinv}
\mathbf{c}^{-1}(\boldsymbol{\zeta}) = \underset{\mathbf{p} \in \mathcal{S}}{\operatorname{argmin}} \left\langle \boldsymbol{\Psi}(\mathbf{p}), \mathbf{V}\boldsymbol{\zeta} \right\rangle_{\mathcal{H}},
\end{equation}
where $\boldsymbol{\zeta} \in \mathcal{H}$.

Finally, by combining \eqref{eq_gt} with \eqref{eq_cinv}, the estimated output can be exposed as~\cite{ciliberto2020general}
\begin{equation}\label{predictor}
\mathbf{s}(t) =  \underset{\mathbf{p} \in \mathcal{S}}{\operatorname{argmin}} \sum_{n=1}^{N}\boldsymbol{\alpha}_n(t) \Delta(\mathbf{p}, \mathbf{p}_n).
\end{equation}
The usage of estimator \eqref{predictor} consists of two steps: 1) \textit{Training}: the score function $\boldsymbol{\alpha}$ is computed given a test input $t$, and 2) \textit{Prediction}: a linear $\alpha_n$-weighted cost is minimized. 
Notably, there is no need for explicit knowledge of the space $\mathcal{H}$, the feature map $\boldsymbol{\Psi}$, or the operator $\mathbf{V}$ thanks to the structure of the loss $\Delta$.
Hence an implicit embedding framework is attained, as shown in Fig.~\ref{fig:illuKerSP}.

\subsection{Kernel Design for Motion Pattern Capture}\label{sec_kernels}
As in other laser-related applications, the trajectory of the cosmetic laser also exhibits the property of (quasi-)periodicity.
To capture such a rhythmic motion pattern, it is not a trivial issue to consider the design of the reproducing kernels since the form of the kernel significantly determines the \textit{generalization} capability of a learning model.

The impact of kernel functions on pattern analysis within nonparametric learning methods has been a crucial concern.
The design of the kernels for pattern analysis has been well studied from the Bayesian perspective such as Gaussian processes (GP) where the kernels appear as covariance functions.
Also, kernel properties are exploited in the frequentist learning field for supervised tasks.
Despite Bayesian learning with GP and frequentist methods with RKHS representing two distinct approaches to non-parametric regression, the properties of the kernels can indeed be harnessed due to the connections between Bayesian and frequentist regularization approaches~\cite{kimeldorf1970correspondence}.

\begin{remark}
Prediction with RKHS as in~\eqref{rkhspre} is equivalent to the mean prediction of a multi-task GP with the zero mean prior and noise variance being $\lambda N \mathbf{I}_{N}$ using the same kernel.
\end{remark}

Given the equivalence between GP and RKHS, the constitution of kernels considered in the context of GP can also be directly leveraged into the regularization framework of RKHS.
In other words, it is reasonable for one to design kernels for RKHS by referring to the properties of the kernels in GP as discussed by~\cite[Ch.~2]{duvenaud2014automatic}.

As a \textit{universal} kernel, the squared exponential kernel is commonly employed as a default kernel in kernel-based machine learning, which has the form: 
\begin{equation}\label{sek}
    k_{\mathtt{SE}}(t, t') = \sigma_s^2\exp\left(- \frac{(t-t')^2}{2l_s^2}\right),
\end{equation} 
where length-scale $l_s$ and variance $\sigma_s^2$ are the hyper-parameters.
Despite its prevalence, the squared exponential kernel could fail to capture certain patterns.
We shall see shortly that it is not suitable for expressing the robot's periodic movement, whose definition is given as follows.

\begin{definition}[Periodicity of a geometric trajectory]
A periodic geometric trajectory is defined by
\begin{equation}\label{riemannianper}
    \mathbf{P}(t) = \mathbf{P}(t+T),
\end{equation}
where $T>0$ is called the period of the geometric trajectory.
\end{definition}

In the following, we will show the necessity of choosing a proper kernel for capturing our motion pattern.

\begin{proposition}[Model misspecification]\label{mismodel}
Given a query point $t\rightarrow +\infty$, 
the output of the estimator \eqref{predictor} with the squared exponential kernel \eqref{sek} remains constant.   
\end{proposition}
\begin{proof}
 For any two training points $t_{n1}$ and $t_{n2}$, we have
\begin{equation}
\lim_{t\rightarrow +\infty} \dfrac{\mathbf{k}_{n1}(t)}{\mathbf{k}_{n2}(t)}  
= \exp\left(\dfrac{(t-t_{n2})^2 - (t-t_{n1})^2}{2l_s^2}\right)   
= 1,
\end{equation}
which implies that $\mathbf{k}$ becomes a vector with all equal elements with the query point $t$ being far from the training set.
As a result, $\boldsymbol{\alpha}$ becomes independent of the query input, leading to the loss function of the $n$-th element scored by a constant.
In particular, the constant weight is given by $\boldsymbol{\alpha}_n = \sum_{i=1}^{N}[(\mathbf{K}+\lambda N \mathbf{I}_N)^{-1}]_{n, i}$.
\end{proof}
Proposition~\ref{mismodel} claims that the estimator \eqref{predictor} with the squared exponential kernel cannot \textit{extrapolate} well on the periodic dataset since the output is simply a constant as the query point moves far away from the training dataset.  
A similar problem could also happen in the case of \textit{interpolation}.
When the interval between two consecutive training points is larger than one period of the periodic trajectory, the prediction along this interval is also non-periodic, which is caused by the non-periodicity of $\boldsymbol{\alpha}$.

In view of the \textit{model misspecification} issue, we propose to employ the periodic kernel to capture the periodic motion pattern. The form of the periodic kernel is given by~\cite{mackay1998introduction}
\begin{equation}\label{perker}
   k_{\mathtt{PER}}(t, t') = \sigma_p^2\exp\left(- \frac{2\sin^2\big(\pi(t-t')/p\big)}{l^2_p} \right), 
\end{equation}
where period $p$, length-scale $l_p$, and variance $\sigma_p^2$ are the kernel hyper-parameters.
\begin{proposition}[Blessing of Abstraction]\label{blessing}
Given a query point $t$, the output of the estimator \eqref{predictor} with the periodic kernel \eqref{perker} remains periodic.     
\end{proposition}
\begin{proof}
Given a training point $t_n$, we have
\begin{equation}
\begin{split}
\mathbf{k}_n(t)  &=  
\sigma_p^2\exp\left(-\frac{2\sin^2\big(\pi(t-t_n)/p\big)}{l_p^2} \right)\\
 &= \sigma_p^2\exp\left(- \frac{2\sin^2\big(\pi(t+p-t_n)/p\big)}{l_p^2} \right)\\
&=  k_{\mathtt{PER}}(t+p, t_n) = \mathbf{k}_n(t+p),
\end{split}
\end{equation}    
which implies that $\mathbf{k}$ is periodic with the period being $p$.
Therefore, the output of \eqref{predictor} becomes also periodic due to the periodicity of $\boldsymbol{\alpha}$. 
\end{proof}

Proposition~\ref{blessing} shows that based on a properly designed kernel function, our structured predictor can perform well regarding out-of-distribution generalization, as depicted in Fig.~\ref{fig:illuKerSP}, since the underlying motion pattern is accounted for by the structure of the kernel function, which gives rise to the \textit{blessing of abstraction}. 

\begin{figure}[t]
\centering
\noindent\includegraphics[width=0.9\columnwidth]{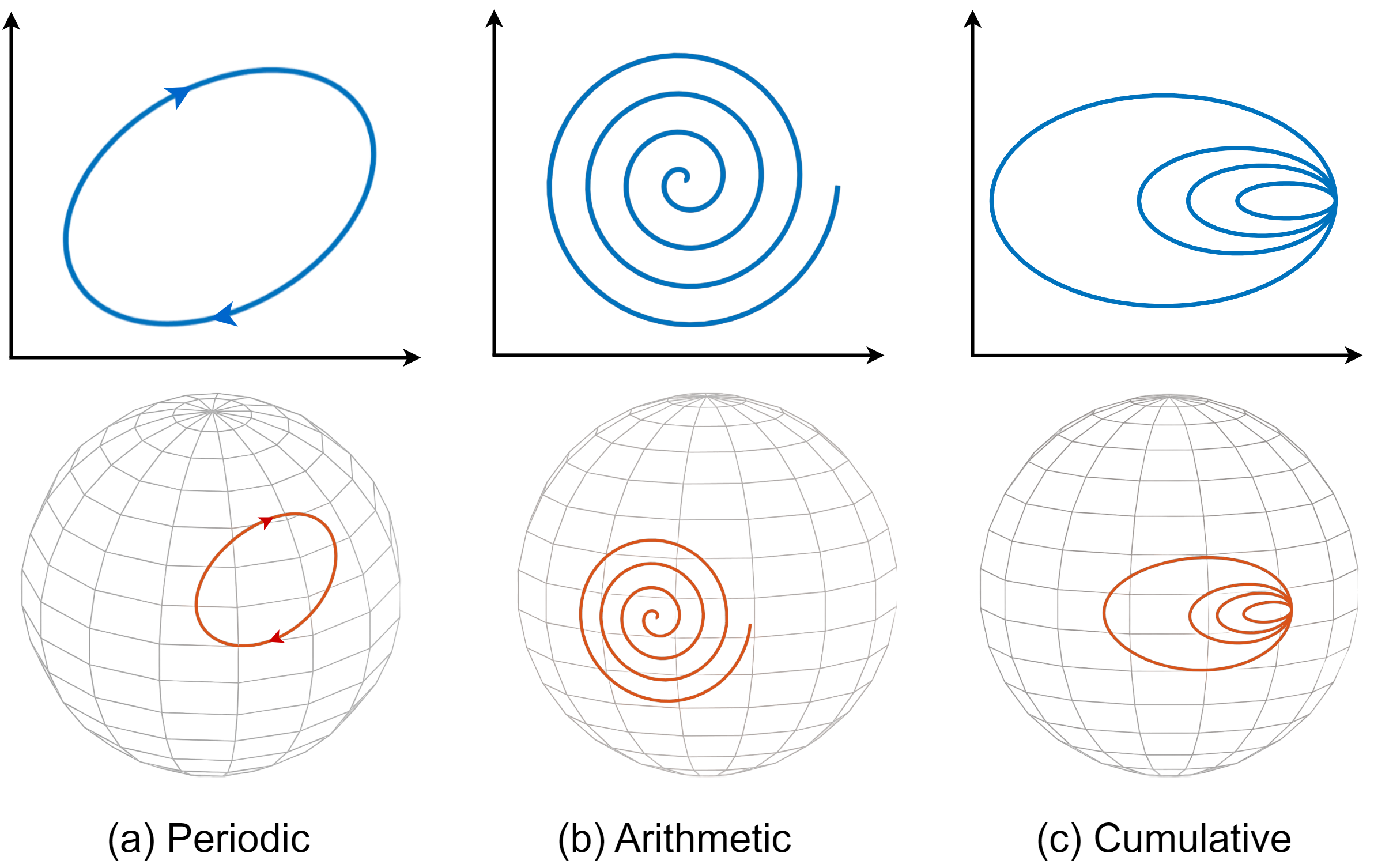}
\caption{Illustration of (a) periodic, (b) arithmetic, and (c) cumulative patterns where Euclidean trajectories (\textit{upper row}) are plotted in blue and trajectories on a manifold (\textit{bottom row}) are plotted in red.}
\label{fig:geo_per_traj}
\end{figure}

It is worth noting that although the motion pattern of the structured output space~\eqref{riemannianper} suggests the usage of the periodic kernel, it is the surrogate space, where the kernel ridge regression is performed as in~\eqref{rkhspre}, that directly prescribes the kernel design.  
Therefore, it is of interest to investigate the behavior of the vectors in the surrogate space.
In fact, given~\eqref{riemannianper} holds, we can conclude that the surrogate space indeed exhibits periodic, namely $\boldsymbol{\Psi}(\mathbf{p}(t)) = \boldsymbol{\Psi}(\mathbf{p}(t+T))$, which can be shown by the bijection property of the feature map.

In practice, the periodic trajectory may not exactly recur itself. 
Most likely, the trajectory could be \textit{quasi-periodic}, i.e. the recurring part may evolve over time~\cite{noorzadeh2014modeling}.
It is therefore necessary to modify the periodic kernel such that the pattern of quasi-periodicity can be captured by the modified kernel. 
In this regard, we propose to employ the quasi-periodic kernel obtained by the product of a periodic kernel and a squared exponential kernel~\cite{angus2018inferring}:
\begin{equation}\label{quasiker}
k_{\mathtt{QP}}(t, t') = k_{\mathtt{SE}}(t, t') \times k_{\mathtt{PER}}(t, t').
\end{equation}

As a sanity check, it is also worth investigating the corresponding behavior of the structured output space given the quasi-periodicity of the surrogate space to justify the usage of the quasi-periodic kernel in the surrogate space learning.  
To this end, we first define two representative quasi-periodic patterns for the case of vector-valued functions, provided $\mathbf{q}(t)$ is defined on $[0, T]$.   

\begin{definition}[Quasi-Periodicity for vector-valued function]\label{defvvqp}
\begin{subequations}
$\mathbf{Arithmetic}:$
\begin{equation}
\mathbf{q}(\tau+aT) = \mathbf{q}(\tau+(a-1)T) + \mathbf{C}(\tau),\label{defvvarith}
\end{equation}    
$\mathbf{Cumulative}:$
\begin{equation}
\mathbf{q}(\tau+aT) = \mathbf{q}(\tau+(a-1)T) + a\mathbf{C}(\tau),\label{defvvcumu}
\end{equation}
\end{subequations}
where $\tau \in [0, T]$, $a \in \mathbb{N}^{+}$ denotes the period index, and $\mathbf{C}(\tau)$ is a continuous function deciding the temporal change following the first period.
\end{definition}

Akin to Definition~\ref{defvvqp}, we extend the corresponding concepts of quasi-periodicity to their geometric counterparts.

\begin{definition}[Quasi-Periodicity for geometric trajectory]
$\mathbf{Arithmetic}:$
\begin{subequations}
\begin{equation}
\mathbf{P}(\tau+aT) = \mathtt{Exp}_{\mathbf{P}(\tau+(a-1)T)}\boldsymbol{\Gamma}_{\mathbf{P}(\tau)\rightarrow\mathbf{P}(\tau+(a-1)T)}\mathfrak{C}(\tau),\label{geoarith} 
\end{equation} 
$\mathbf{Cumulative}:$
\begin{equation}
\mathbf{P}(\tau+aT) = \mathtt{Exp}_{\mathbf{P}(\tau+(a-1)T)}\boldsymbol{\Gamma}_{\mathbf{P}(\tau)\rightarrow\mathbf{P}(\tau+(a-1)T)}a\mathfrak{C}(\tau),\label{geocumu}
\end{equation}
\end{subequations}
where $\mathbf{P}(t)$ is given on $[0, T]$, and $\mathfrak{C}(\tau)$ is defined on the tangent space at $\mathbf{P}(\tau)$.
\end{definition}

Fig.~\ref{fig:geo_per_traj} illustrates the periodic and quasi-periodic behaviors on a plane and a sphere manifold.

\begin{theorem}\label{Quasi_periodic}
If the trajectories in surrogate space satisfy quasi-periodic behavior as in     
\eqref{defvvarith}
or
\eqref{defvvcumu},
then the geometric trajectories conform to
\eqref{geoarith}
and
\eqref{geocumu}, respectively.
\end{theorem}

\begin{proof}
See Appendix \ref{app_quasi}.
\end{proof}

Remarkably, Theorem~\ref{Quasi_periodic} states that the quasi-periodic pattern in the surrogate space denotes a sufficient condition for the geometric output space to be quasi-periodic as well.

\subsection{Loss Function Calculation}\label{sec_loss}       
When applying \eqref{predictor} for learning geometric trajectories, we need to solve a geometric optimization problem that involves the computation of geodesic distances. 
The geodesic path on 3D surfaces is usually difficult to find and the procedure may require many iterations~\cite{bose2011survey}.
Provided that we need to compute the geodesic distance of a manifold point to all demonstrated trajectory points, it is thus favorable to efficiently compute the geodesic distance of two points on the facial surface.   

We propose to exploit the strategy of \textit{spherelets}, which refers to manifold approximation using spheres~\cite{10.1111/rssb.12508}. 
Geodesic distance calculation based on the spherelet technique is efficient as it admits a closed-form solution. 
The main steps of using spherelets for geodesic distance calculation are sketched as follows: 1) Partitioning the facial surface into composing regions, 2) Designating an \textit{osculating sphere} for each region to approximate the manifold, and 3) Calculating the geodesic distance by summing the length of each composing geodesic path segment.

\begin{figure}[t]
    \centering
    \includegraphics[width=0.86\columnwidth]{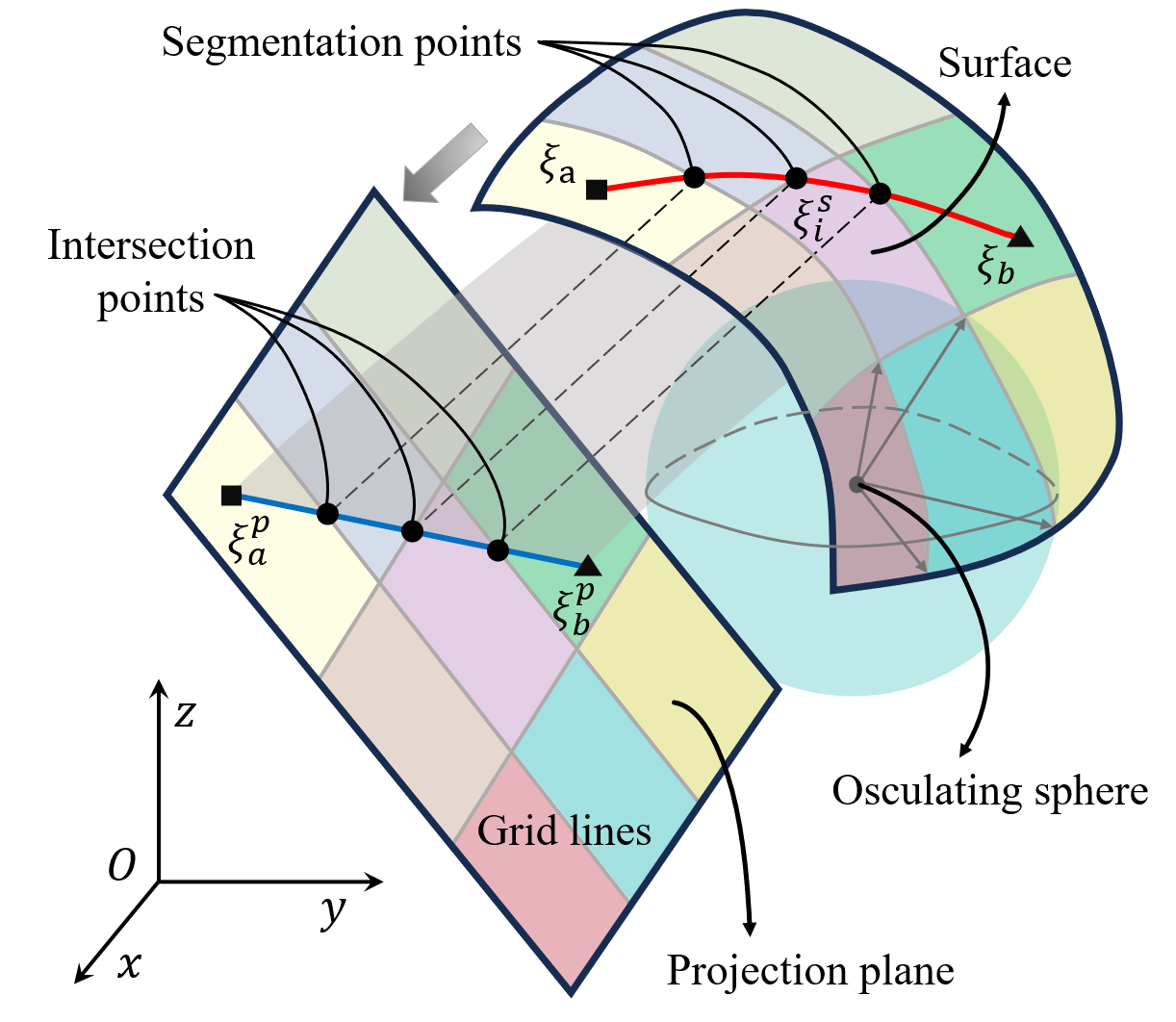}
    \caption{Illustration of geodesic distance calculation by spherelets. The geodesic distance between two points as denoted by a square and a triangle on the surface is obtained by summing the lengths of the geodesic curves on the composing spheres. The segmentation points are determined using the intersection points between the mapped straight curve and the specified grid lines on the projection plane.}
    \label{fig:geo_dist_cal}
\end{figure}

We consider partitioning the facial surface by a grid.
As a result, the surface manifold is composed by $\mathcal{S} = \bigcup_{s=1}^{d} \mathbb{S}_s^2$, where $d$ denotes the total number of regions. 
As mentioned earlier, we propose to use an osculating sphere to approximate each surface region.
Formally, an osculating sphere is defined by 
$\mathbb{S}_s^2 := \{\boldsymbol{\zeta}: \|\boldsymbol{\zeta} - \mathbf{O}_s\|=r_s \}$.
Given the 3D points collection of the facial surface 
$\mathbb{F}: \{\boldsymbol{\xi}_m = (x_m, y_m, z_m) \}_{m=1}^{M}$, the center $\mathbf{O}_s$ and the radius $r_s$ are obtained by minimizing the following \textit{algebraic} loss function:
\begin{equation}\label{algebraic_loss}
C_s(\mathbf{O}_s, r_s)= \sum_{\boldsymbol{\xi}_{i} \in \mathbb{S}_s^2}^{}
\left((\boldsymbol{\xi}_{i}-\mathbf{O}_s)^\intercal(\boldsymbol{\xi}_{i}-\mathbf{O}_s) - r_s^2\right)^2. 
\end{equation}
\begin{lemma}[Osculating Sphere Estimation \cite{10.1111/rssb.12508}]\label{lemma_sphereminimizer}
The minimizer to \eqref{algebraic_loss} is given by
\begin{equation}\label{sphereminimizer}
\mathbf{O}_s = \dfrac{1}{2}\boldsymbol{\Lambda}_s^{-1}\boldsymbol{\theta}_s    
\quad \text{and} \quad
r_s = \dfrac{1}{M_s} \sum_{\boldsymbol{\xi}_{i} \in \mathbb{S}_s^2}^{}\|\boldsymbol{\xi}_i - \mathbf{O}_s\|,     
\end{equation}
where $M_s$ denotes the number of data points in the corresponding region and
\begin{subequations}
\begin{equation}
\boldsymbol{\Lambda}_s=\sum_{\boldsymbol{\xi}_{i} \in \mathbb{S}_s^2}^{}(\boldsymbol{\xi}_{i}-\overline{\boldsymbol{\xi}}_s)(\boldsymbol{\xi}_{i}-\overline{\boldsymbol{\xi}}_s)^\intercal,     
\end{equation}
\begin{equation}
\boldsymbol{\theta}_s=\sum_{\boldsymbol{\xi}_{i} \in \mathbb{S}_s^2}^{}(\boldsymbol{\xi}_{i}^\intercal\boldsymbol{\xi}_{i}- 
\dfrac{1}{M_s}\sum_{\boldsymbol{\xi}_{i} \in \mathbb{S}_s^2}^{}\boldsymbol{\xi}_{i}^\intercal\boldsymbol{\xi}_{i})(\boldsymbol{\xi}_{i}-\overline{\boldsymbol{\xi}}_s). 
\end{equation}
\end{subequations}
\end{lemma}

\begin{algorithm}[t]
	\caption{Imitation of Geometric Trajectory with Rhythm by Structured Prediction} 
	\label{permanifold_alg}
    Scan the subject's face for point cloud dataset $\mathbb{F}$\;
    Determine the projection plane perpendicular to the eigenvector with the minimum eigenvalue of \eqref{facedatacov}\;
    Specify grid $\{\{\boldsymbol{l}^p_{vi}\}_{i=1}^{nv}, \{\boldsymbol{l}^p_{hj}\}_{j=1}^{nh}\}$ for facial partition\;
    Compute the centers and radius as per \eqref{sphereminimizer}\; 
	Collect the demonstrated trajectory for dataset $\mathbb{T}$\;
	Define the kernel function $k(\cdot, \cdot)$ and hyperparameters\;
	Choose the regularization term $\lambda$ and the step size $\eta$\;
	\For{$t = t_{\mathtt{init}}, \ldots, t_{\mathtt{end}}$}{
		\textit{Input:} Query point $t$\;
		Calculate the scores $\boldsymbol{\alpha}(t)$\;
        \Repeat{convergence}{
            \For{$n = 1, \ldots, N$}{
            Determine the intersection points with the grid lines on the projection plane\;
            Map the intersection points to the surface manifold for segmentation points\;
            Calculate the geodesic distance to each training point $\mathcal{D}(\mathbf{P}(t), \mathbf{P}_n)$ as per \eqref{geo_dist_approx}\;
            }
            Compute the Riemannian gradient as per \eqref{Riemanniangrad}\;
            Update the prediction as per \eqref{itrexpmap}\;   
		}	
		\textit{Output:} Prediction $\mathbf{P}(t)$\;
	}
\end{algorithm}

To determine the segmentation points on the geodesic path connecting two manifold points of interest $\boldsymbol{\xi}_a$ and $\boldsymbol{\xi}_b$, we identify a plane that is determined by the maximum variance of the facial surface, i.e. the projected area of the face surface onto the plane is maximized.
To this end, we compute the data covariance matrix as
\begin{equation}\label{facedatacov}
\boldsymbol{\Sigma}_f = \frac{1}{M}\!\sum_{m=1}^{M}(\boldsymbol{\xi}_m-\overline{\boldsymbol{\xi}})(\boldsymbol{\xi}_m-\overline{\boldsymbol{\xi}})^\intercal, \, \text{with} \;\, \overline{\boldsymbol{\xi}} = \frac{1}{M}\!\sum_{m=1}^{M}\boldsymbol{\xi}_m.
\end{equation}
The projection plane is then identified by aligning its normal with the least principal component of the data points, which is the eigenvector with the smallest eigenvalue of $\boldsymbol{\Sigma}_f$.
By projecting $\boldsymbol{\xi}_a$ and $\boldsymbol{\xi}_b$ to the projection plane, we obtain the projected points $\boldsymbol{\xi}_a^p$ and $\boldsymbol{\xi}_b^p$, respectively. 
On the projection plane, we compute the intersection points by crossing the straight line $\overline{\boldsymbol{\xi}_a^p\boldsymbol{\xi}_b^p}$ with the grid lines $\{\{\boldsymbol{l}^p_{vi}\}_{i=1}^{nv}, \{\boldsymbol{l}^p_{hj}\}_{j=1}^{nh}\}$ where $\boldsymbol{l}^p_{vi}$ denote the vertical lines and $\boldsymbol{l}^p_{hj}$ denote the horizontal lines.
Then, the segmentation points are obtained by projecting the intersection points on the projection plane to the approximated manifold, i.e., 
$\boldsymbol{\xi}^s_{vi, hi} = 
\boldsymbol{\Pi}_p^s(\boldsymbol{l}^p_{vi}, \boldsymbol{l}^p_{hi} \cap\, \overline{\boldsymbol{\xi}_a^p\boldsymbol{\xi}_b^p})$ where $\boldsymbol{\Pi}_p^s(\cdot)$ perpendicularly maps a point on the projection plane to the point lying on the corresponding approximated sphere manifold. 

Finally, by sorting the segmentation points to be 
$\{\boldsymbol{\xi}^s_{i}\}_{i=1}^{L}$ such that they are ordered in accordance with the direction of $\boldsymbol{\xi}_a^p$ to $\boldsymbol{\xi}_b^p$, the geodesic distance can be approximated by
\begin{equation}\label{geo_dist_approx}
\mathcal{D}(\boldsymbol{\xi}_a, \boldsymbol{\xi}_b) \approx \sum_{l=0}^{L}r_l \arccos\left(\frac{\boldsymbol{\xi}^s_{l}-\mathbf{O}_l}{r_l} \cdot \frac{\boldsymbol{\xi}^s_{l+1} - \mathbf{O}_l}{r_l} \right),
\end{equation}
where $r_l$ and $\mathbf{O}_l$ represent the radius and center of the corresponding sphere, respectively. 
Besides, we have $\boldsymbol{\xi}_0 = \boldsymbol{\xi}_a$ and $\boldsymbol{\xi}_{L+1} = \boldsymbol{\xi}_b$.
Fig.~\ref{fig:geo_dist_cal} provides a pictorial illustration of geodesic distance approximation using spherelets.

Given the solution to the geodesic distance calculation by \eqref{geo_dist_approx}, the geometric optimization problem in \eqref{predictor} can be readily addressed by \textit{Riemannian gradient descent}, which represents the Riemannian counterpart of the usual gradient descent method~\cite{ciliberto2020general}.
The update rule iteratively proceeds as
\begin{equation}\label{itrexpmap}
\mathbf{p}_{i+1} = \mathtt{Exp}_{\mathbf{p}_i}\left(\eta_i \nabla_{\mathcal{M}}\mathtt{F}(\mathbf{p}_i)\right),
\end{equation}
where $\eta_i \in \mathbb{R}$ is a step size, and we denote 
\begin{subequations}
\begin{equation}
\mathtt{F}(\mathbf{p}_i) = \sum_{n=1}^{N}\boldsymbol{\alpha}_n(t) \Delta(\mathbf{p}_i, \mathbf{p}_n),
\end{equation}
\begin{equation}
\Delta(\mathbf{p}_i, \mathbf{p}_n) 
= \mathcal{D}^2(\mathbf{p}_i, \mathbf{p}_n).   
\end{equation}
\end{subequations}
Specifically, $\nabla_{\mathcal{M}}$ denotes the Riemannian gradient operator and the gradient defined with respect to the Riemannian metric is given by the projection of the usual gradient to the tangent space~\cite{absil2008optimization}, which reads
\begin{equation}\label{Riemanniangrad}
\nabla_{\mathcal{M}}\mathtt{F}(\mathbf{p}_i) = \mathtt{Proj}_{\mathbf{p}_i}\nabla\mathtt{F}(\mathbf{p}_i).
\end{equation}
Wherein we have
\begin{subequations}
\begin{flalign}
&\text{\textit{Orthogonal projector}:} \quad \mathtt{Proj}_{\mathbf{p}_i} \triangleq  r_i^2\mathbf{I} - \mathbf{p}_i\mathbf{p}_i^\intercal,&   
\end{flalign}
\begin{flalign}
&\text{\textit{Euclidean gradient}:} \quad \nabla\mathtt{F}(\mathbf{p}_i) \triangleq  2\sum_{n=1}^{N}\boldsymbol{\alpha}_n \mathcal{D}_{i, n}\dfrac{\partial \mathcal{D}_{i, n}}{\partial \mathbf{p}_i}.& 
\end{flalign}
\end{subequations}
We denote $\mathcal{D}_{i, n} := \mathcal{D}(\mathbf{p}_i, \mathbf{p}_n)$ for simplicity.

The algorithm for imitating rhythmic trajectory on a facial surface using structured prediction is summarized in Alg.~\ref{permanifold_alg}.

\section{Adaptation of Learned Skills \\to Treat New Faces}\label{adaptation}
As a central topic in robot learning from demonstration, the adaptation of motor skills is crucial for robots to accomplish the task in an environment that is different from the one in the demonstration.  
In our application, the objective is to equip the robot with the capacity to execute treatment on subjects who were not present during the demonstration.   
To achieve adaptation of treatment skills, our strategy is to modulate trajectory in accordance with changes in the shape of human faces. 
Wherein, human facial shapes are captured by key facial features such as control points.
Consequently, based on changes in the key facial features, we acquire rules for the laser generator's trajectory adaptation. 

We propose to tackle the problem of discovering the rules for trajectory adaptation by \textit{feature-based registration}, whose goal is to align a source surface (i.e., the face in the demonstration) with a target surface (i.e., the face during the adaptation) given landmark points.  
Specifically, we consider the \textit{nonrigid} transformation, which permits soft deformations of the point sets. 
A common technique for non-rigid registration is to parametrize the transformation by leveraging \textit{thin-plate spline}~\cite{wahba1990spline}.
While nonrigid registration has gained attention in robot imitation learning for modulating learned trajectories towards new environments, the applications have been limited to deterministic scenarios~\cite{schulman2016learning}.

\begin{figure}[t]
    \centering
    \includegraphics[width=0.96\columnwidth]{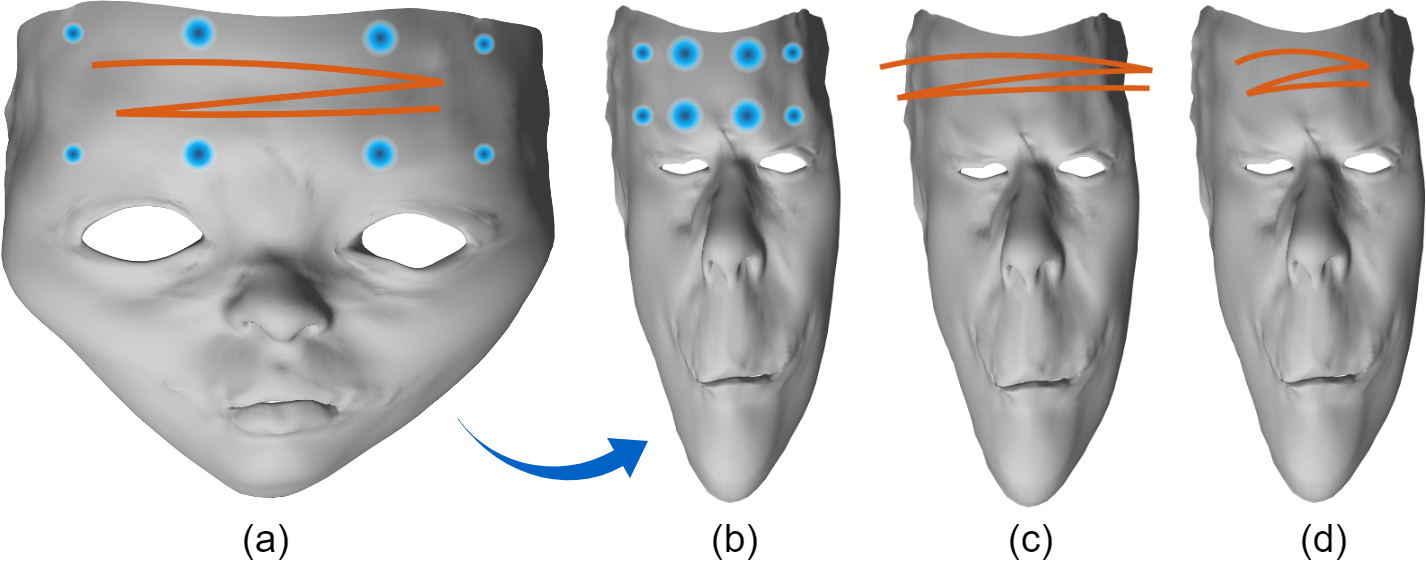}
    \caption{Illustration of probabilistic nonrigid registration. (a) The original shape of the trajectory on the demonstration face. (b) The adaptation face where control points are depicted in blue with the radius indicating the covariance matrix contour. (c) Transformed trajectory shape without probabilistic registration, which exceeds the boundary of the forehead. (d) Transformed trajectory shape with probabilistic registration which ensures that the trajectory lies within the boundary of the forehead.}
    \label{fig:prob_reg}
\end{figure}

In the following,  we present a method for achieving \textit{probabilistic} trajectory adaptation with nonrigid registration.
By incorporating additional probabilistic information in the form of associated covariance matrices, our method offers greater flexibility in devising the transformation.
Specifically, trajectory points with higher variances will be allowed to deviate more while those with lower variances will deflect less after the transformation. 
The proposed probabilistic trajectory adaptation technique is particularly relevant to our application of skincare, where covariance matrices could be utilized to encode local skin conditions.

Formally, assume that there are $K$ landmark points for the source and target facial surface which are represented by
$\{\hat{\boldsymbol{\xi}}_{l_i}\}_{i=1}^{K}$ and
$\{\hat{\boldsymbol{\xi}}_i'\}_{i=1}^K$, respectively,
with the points index stored as $\{l_1,\ldots,l_K\}$.
In addition, the associated covariance matrices are given by 
$\{\hat{\boldsymbol{\Sigma}}_{l_i}\}_{i=1}^{K}$.
Our goal is to find a transformation rule $\boldsymbol{f}: \mathbb{R}^D \rightarrow \mathbb{R}^D$ such that each original source point can be mapped to its corresponding target point.
The problem of determining the transformation function $\boldsymbol{f}$ can be cast as the following optimization problem:
\begin{equation}\label{eq:tpsopt}
	E(\boldsymbol{f}) \!=\! \sum_{i=1}^{K}\big(\hat{\boldsymbol{\xi}}_i' \!-\! \boldsymbol{f}(\hat{\boldsymbol{\xi}}_{l_i})\big)^\intercal \hat{\boldsymbol{\Sigma}}_{l_i}^{-1} \big(\hat{\boldsymbol{\xi}}_i' \!-\! \boldsymbol{f}(\hat{\boldsymbol{\xi}}_{l_i})\big) \!+\! \lambda \|\boldsymbol{f}\|^2_{\mathtt{tps}}
\end{equation}
where $E(\cdot)$ is the so-called bending energy.
As a slight abuse of notation, $\lambda>0$ is a smoothing parameter that balances the trade-off between smoothness and goodness-of-fit.
Besides, $\|\boldsymbol{f}\|^2_{\mathtt{tps}}$ denotes the thin plate spline regularization term,  which imposes a soft constraint on smoothness in order to control the behavior of the mapping.
It is defined by the space integral of the square of the second-order derivatives~\cite{wahba1990spline}:
\begin{equation}
	\label{eq:reg}
	\|\boldsymbol{f}\|^2_{\mathtt{tps}} = \int_{\mathbb{R}^3}
    \left\| \dfrac{\partial^2 \boldsymbol{f}}{\partial^2 \hat{\boldsymbol{\xi}}} \right\|_{\mathtt{Frob}}^2 d \hat{\boldsymbol{\xi}}.
\end{equation}

Remarkably, there exists a unique minimizer, and the analytical solution to $\boldsymbol{f}$ is composed of an affine or rigid part and a nonlinear or nonrigid part: 
\begin{equation}\label{tpssol}
	\boldsymbol{f}(\hat{\boldsymbol{\xi}}_{l_i}) =
	\underbrace{\mathbf{B}\hat{\boldsymbol{\xi}}_{l_i}}_\text{Affine part} + \underbrace{\boldsymbol{\omega}\boldsymbol{\rho}(\hat{\boldsymbol{\xi}}_{l_i})}_\text{ Nonlinear term}, 
\end{equation}
where $\mathbf{B} \in \mathbb{R}^{(D+1)\times(D+1)}$ accounts for the affine transformation and $\boldsymbol{\omega} \in \mathbb{R}^{(D+1) \times K}$ contains the warping coefficients enabling non-affine deformation.
It should be noted that we augment each point with an intercept term for homogeneous coordinates, i.e., 
$ \hat{\boldsymbol{\xi}}_{l_i} \leftarrow 
[\hat{\boldsymbol{\xi}}_{l_i}^\intercal \; 1]^\intercal$
in order to include data offset conveniently.
The vector $\boldsymbol{\rho}(\hat{\boldsymbol{\xi}}_{l_i}) \in \mathbb{R}^{K}$ is composed of the thin-plate spline basis functions and each entry of the vector is defined by 
\begin{equation}
\boldsymbol{\rho}_j(\hat{\boldsymbol{\xi}}_{l_i}) = - \left\|\hat{\boldsymbol{\xi}}_{l_j} - \hat{\boldsymbol{\xi}}_{l_i} \right\|^2.
\end{equation}

To solve for $\mathbf{B}$ and $\boldsymbol{\omega}$, we substitute \eqref{tpssol} into \eqref{eq:tpsopt}.
Then, the bending energy becomes 
\begin{align}
		E &=  \mathtt{sum}\big((\widetilde{\boldsymbol{\xi}}'-\mathbf{G}\boldsymbol{\omega}^\intercal-\widetilde{\boldsymbol{\xi}}_l\mathbf{B}^\intercal)\odot\widetilde{\mathbf{Q}}\odot(\widetilde{\boldsymbol{\xi}}'-\mathbf{G}\boldsymbol{\omega}^\intercal-\widetilde{\boldsymbol{\xi}}_l\mathbf{B}^\intercal)\big)& \nonumber\\ 
        & \qquad \quad + \lambda \mathtt{Tr}(\boldsymbol{\omega} \mathbf{G} \boldsymbol{\omega}^\intercal) \nonumber\\
		&=  \| \widetilde{\mathbf{Q}}^{\frac{1}{2}}\odot (\widetilde{\boldsymbol{\xi}}'\!-\!\mathbf{G}\boldsymbol{\omega}^\intercal \!-\!\widetilde{\boldsymbol{\xi}}_l\mathbf{B}^\intercal)\|_{\mathtt{Frob}}^2 \!+\! \lambda \mathtt{Tr}(\boldsymbol{\omega} \mathbf{G} \boldsymbol{\omega}^\intercal)
\end{align}
where $\mathbf{G} \in \mathbb{R}^{K \times K}$ is constructed by 
$\mathbf{G}_{i, j} =\boldsymbol{\rho}_j(\hat{\boldsymbol{\xi}}_{l_i})$
and we have the concatenated terms:
\begin{align}\label{eq:concaten}
\begin{split}
&\widetilde{\boldsymbol{\xi}}_l = 
\begin{bmatrix}
\hat{\boldsymbol{\xi}}_{l_1}, \ldots, \hat{\boldsymbol{\xi}}_{l_K}
\end{bmatrix}^\intercal,
\quad
\widetilde{\boldsymbol{\xi}}' = 
\begin{bmatrix}
\hat{\boldsymbol{\xi}}_1',\ldots,\hat{\boldsymbol{\xi}}_K'
\end{bmatrix}^\intercal,\\[0.6em]
&\text{and} \quad 
\widetilde{\mathbf{Q}} = 
\mathtt{blockdiag} 
(\hat{\boldsymbol{\Sigma}}_{l_1}^{-1},
\ldots,
\hat{\boldsymbol{\Sigma}}_{l_K}^{-1}).
\end{split}
\end{align}

In view of the optimization objective, it is not straightforward to calculate $\mathbf{B}$ and $\boldsymbol{\omega}$.
Therefore, the QR decomposition is employed to separate the affine and non-affine warping spaces~\cite{wahba1990spline}.
Specifically, we have
\begin{equation}\label{QRdecomposition}
	\widetilde{\mathbf{Q}}^{\frac{1}{2}} \odot \widetilde{\boldsymbol{\xi}}_l = [\mathbf{Q}_1 \   \mathbf{Q}_2]\begin{bmatrix}
		\mathbf{R} \\
		\mathbf{0}
	\end{bmatrix}.
\end{equation} 
By observing that $(\widetilde{\mathbf{Q}}^{\frac{1}{2}} \odot \widetilde{\boldsymbol{\xi}}_l)^\intercal \boldsymbol{\omega}^\intercal = \mathbf{0}$, we set
\begin{equation}\label{eq:expressionw}
\boldsymbol{\omega}^\intercal = \mathbf{Q}_2\boldsymbol{\gamma}.    
\end{equation}
Subsequently, the bending energy can be expressed as
\begin{align}
	\begin{split}
		E = &\|\mathbf{Q}_2^\intercal\big( \widetilde{\mathbf{Q}}^{\frac{1}{2}} \odot(\widetilde{\boldsymbol{\xi}}' - \mathbf{G} \mathbf{Q}_2\boldsymbol{\gamma})\big)  \|_{\mathtt{Frob}}^2 \\ &+ 
		\|\mathbf{Q}_1^\intercal\big( \widetilde{\mathbf{Q}}^{\frac{1}{2}}\odot (\widetilde{\boldsymbol{\xi}}'-\mathbf{G}\mathbf{Q}_2\boldsymbol{\gamma})-\mathbf{R}\mathbf{B}^\intercal \big) \|_{\mathtt{Frob}}^2 \\ &+ 
		\lambda \mathtt{Tr}(\boldsymbol{\gamma}^\intercal\mathbf{Q}_2^\intercal \mathbf{G} \mathbf{Q}_2\boldsymbol{\gamma}).
	\end{split}
\end{align}
Finally, the solution for $\mathbf{B}$ is given by
\begin{equation}\label{sol_to_d}
\mathbf{B}^\intercal = \mathbf{R}^{-1}\mathbf{Q}_1^\intercal\big( \widetilde{\mathbf{Q}}^{\frac{1}{2}}\odot (\widetilde{\boldsymbol{\xi}}'-\mathbf{G}\mathbf{Q}_2\boldsymbol{\gamma})\big),
\end{equation}
where we have
\begin{equation}\label{eq:gamma}
\boldsymbol{\gamma} = \big(\mathbf{Q}_2^\intercal\widetilde{\mathbf{Q}}^{\frac{1}{2}}\odot\mathbf{G}\mathbf{Q}_2\!+\!\lambda\mathbf{I} \big)^{\!-1}\mathbf{Q}_2^\intercal\widetilde{\mathbf{Q}}^{\frac{1}{2}}\odot\widetilde{\boldsymbol{\xi}}'. 
\end{equation}

Moreover, the expression for $\boldsymbol{\omega}$ can be readily obtained by combining \eqref{eq:expressionw} and \eqref{eq:gamma}.

\begin{algorithm}[t]
	\caption{Skill Adaptation toward Treating Novel Faces by Probabilistic Nonrigid Registration} 
	\label{skill_adapt}
	Choose source control points $\{\hat{\boldsymbol{\xi}}_{l_i}\}_{i=1}^{K}$\;
	Scan the new subject's face for the point cloud dataset\;
	Determine target control points $\{\hat{\boldsymbol{\xi}}_i'\}_{i=1}^K$\; 
	Specify the associated covariance matrices $\{\hat{\boldsymbol{\Sigma}}_{l_i}\}_{i=1}^{K}$\;
    Concatenate terms for $\widetilde{\boldsymbol{\xi}}_l$, $\widetilde{\boldsymbol{\xi}}'$, and $\widetilde{\mathbf{Q}}$ as per \eqref{eq:concaten}\;
    Perform QR decomposition as per \eqref{QRdecomposition}\;
    Solve for the affine transformation $\mathbf{B}$ and nonlinear warping coefficients $\boldsymbol{\omega}$ as per \eqref{sol_to_d} and \eqref{eq:expressionw}\;
    \textit{Input:} Demonstrated trajectory point $\mathbf{P}(t)$\;
    \textit{Output:} Adaptation point $\boldsymbol{f}(\mathbf{P}(t))$ as per \eqref{tpssol}\;
\end{algorithm}

The effect of probabilistic non-rigid registration is shown in Fig.~\ref{fig:prob_reg}, where, as an illustration, we use two caricature faces~\cite{cai2021landmark}.
The additional regulation can ensure that the adapted trajectory upon non-rigid registration can properly evolve on the new subject's face. 

The steps for the adaptation of laser shot traces towards treating a new face are summarized in Alg.~\ref{skill_adapt}.

\section{Results}\label{experiments}
In this section, we present the results of numerical and experimental studies.
Firstly, we showcase the effectiveness of kernel functions in capturing motion patterns (Section \ref{sec_kereff}), 
the spherelet technique in calculating the geodesic distances (Section \ref{sec_geodist}), 
and non-rigid registration in adapting treatment trajectories (Section \ref{sec_skilladapt}). 
Then, We provide a comparison with state-of-the-art methods for learning periodic geometry-structured trajectories in Section \ref{sec_compsota}.
Finally, in Section \ref{sec_realexp}, we conduct real-world experiments on robotic cosmetic dermatology using photorejuvenation.

\begin{table}[t]
\small
\aboverulesep=0ex 
\belowrulesep=0ex 
\renewcommand{\arraystretch}{1.2}
\caption{Imitation Errors of Reproduction and Generalization}
\centering
\begin{tabular}{cc|c|c|c|c|c}
\toprule
\multicolumn{1}{c}{} & \multicolumn{2}{c}{\textbf{$C$-shape}} & \multicolumn{2}{c}{\textbf{Infinity}} & \multicolumn{2}{c}{\textbf{Spiral}} \\
\cmidrule(rl){2-3} \cmidrule(rl){4-5} \cmidrule(rl){6-7}
($1\mathrm{e}{-3}$) & \multicolumn{1}{c|}{Tra.} & \multicolumn{1}{c}{Gen.}& \multicolumn{1}{c|}{Tra.} & \multicolumn{1}{c}{Gen.}& \multicolumn{1}{c|}{Tra.} & \multicolumn{1}{c}{Gen.}   \\
\hline
SE  & $\mathbf{25.0}$ &  $\mathbf{76.3}$ & $14.1$ & $1258.5$ & $12.3$ & $918.5$ \\
PER & $27.3$ & $217.7$ &  $\mathbf{3.4}$ &    $\mathbf{3.6}$ & $63.5$ & $189.1$ \\
QP  & $25.4$ & $162.4$ &  $5.7$ &  $293.9$ &  $\mathbf{2.9}$ & $\mathbf{37.1}$ \\
\hline
\end{tabular}
\label{Tab_testgeneral}
\end{table}

\begin{figure*}[htp!]
    \centering
    \includegraphics[width=0.96\textwidth]{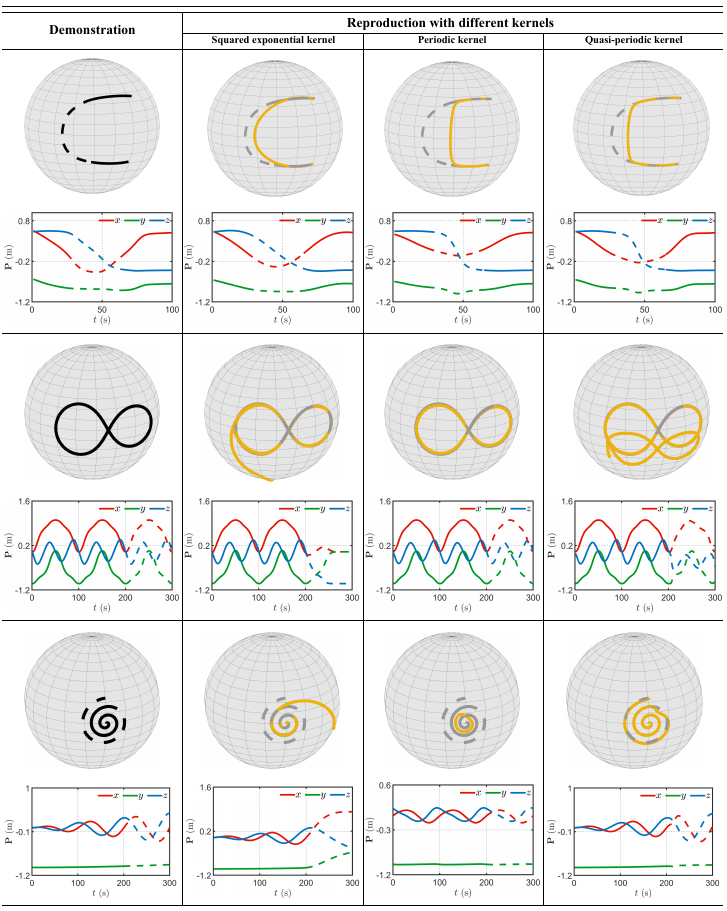}
    \caption{Illustration of learning different motion patterns on a manifold with different kernel functions, where the demonstration trajectory is shown in the \textit{first column}, the motion skills reproduction is achieved by the SE kernel (\textit{second column}), the PER kernel (\textit{third column}), and the QP kernel (\textit{fourth column}). 
    The representative motion patterns to capture are stroke-based trajectory (\textit{top row}), periodic trajectory (\textit{middle row}), and quasi-periodic trajectory (\textit{bottom row}), where the solid lines denote the reproduction part and the dashed lines denote the generalization part.
    For each imitation instance, we first plot the trajectory evolution on the manifold, underneath which we show the temporal evolution of each trajectory dimension.}
    \label{fig:ExpKerEff}
\end{figure*}

\subsection{Effects of Kernel Functions}\label{sec_kereff}
To evaluate the learning effects when using different kernel functions for the estimator \eqref{predictor}, we construct the $\boldsymbol{\alpha}$ scores based on several representative kernel functions to imitate different types of demonstration trajectories, including the Squared Exponential kernel (SE) $k_{\mathtt{SE}}$ as in \eqref{sek}, the Periodic kernel (PER) $k_{\mathtt{PER}}$ as in \eqref{perker}, and the Quasi-Periodic kernel (QP) $k_{\mathtt{QP}}$ as in \eqref{quasiker}.
Three typical patterns of demonstration trajectory, namely stroke-based trajectory, periodic trajectory, and quasi-periodic trajectory are chosen.
These trajectories are designed to evolve on a unit sphere whose center coincides with the origin of the coordinate.
In each learning scenario, we split the whole trajectory into two parts to perform both movement primitive training and skill generalization testing.

We specify a $C$-shaped, an infinity sign $\infty$-shaped, and a spiral-shaped trajectory for stroke-based, periodic, and quasi-periodic trajectory imitation, respectively. 
In the case of learning with the SE kernel, we set $\sigma_s = 5$ and $l_s = 20$.
When using the PER kernel, we set $\sigma_p = 1$, $l_p = 0.5$, and $p = 150$.
As for the QP kernel, the parameters for the composing SE and PER kernels are the same as those used in the standalone cases. 
Besides, the regularization parameter is set to be $\lambda = 0.01$.

The learning results are shown in Fig.~\ref{fig:ExpKerEff}.
It can be observed that the performance of each kernel in different learning situations differs, implying the necessity of a proper design of kernel functions to synthesize the $\boldsymbol{\alpha}$ scores given a specific imitation task. 
For quantitative assessment, we define the following imitation metric based on the average of the accumulation of prediction errors.
To evaluate the reproduction behavior, we have
\begin{equation}\label{eq_evalreprometric}
\mathcal{C}_{\mathrm{T}} = \dfrac{1}{N} \sum_{n=1}^{N} \Delta(\mathbf{s}(t_n), \mathbf{p}(t_n)),
\end{equation}
where $\Delta(\mathbf{s}(t_n), \mathbf{p}(t_n)) = \arccos(\mathbf{s}(t_n)^\intercal\mathbf{p}(t_n))$ in the case of spherical distance.
Similarly, the evaluation metric for examining the generalization effects is given by 
\begin{equation}\label{eq_evalgenmetric}
\mathcal{C}_{\mathrm{G}} = \dfrac{1}{M-N} \sum_{m=N\!+\!1}^{M} \Delta(\mathbf{s}(t_m), \mathbf{p}(t_m)),
\end{equation}
where $M$ represents the total number of prediction steps during generalization. 
The obtained numerical results are summarized in Table~\ref{Tab_testgeneral}.

\begin{figure}[t]
    \centering
    \includegraphics[width=0.99\columnwidth]{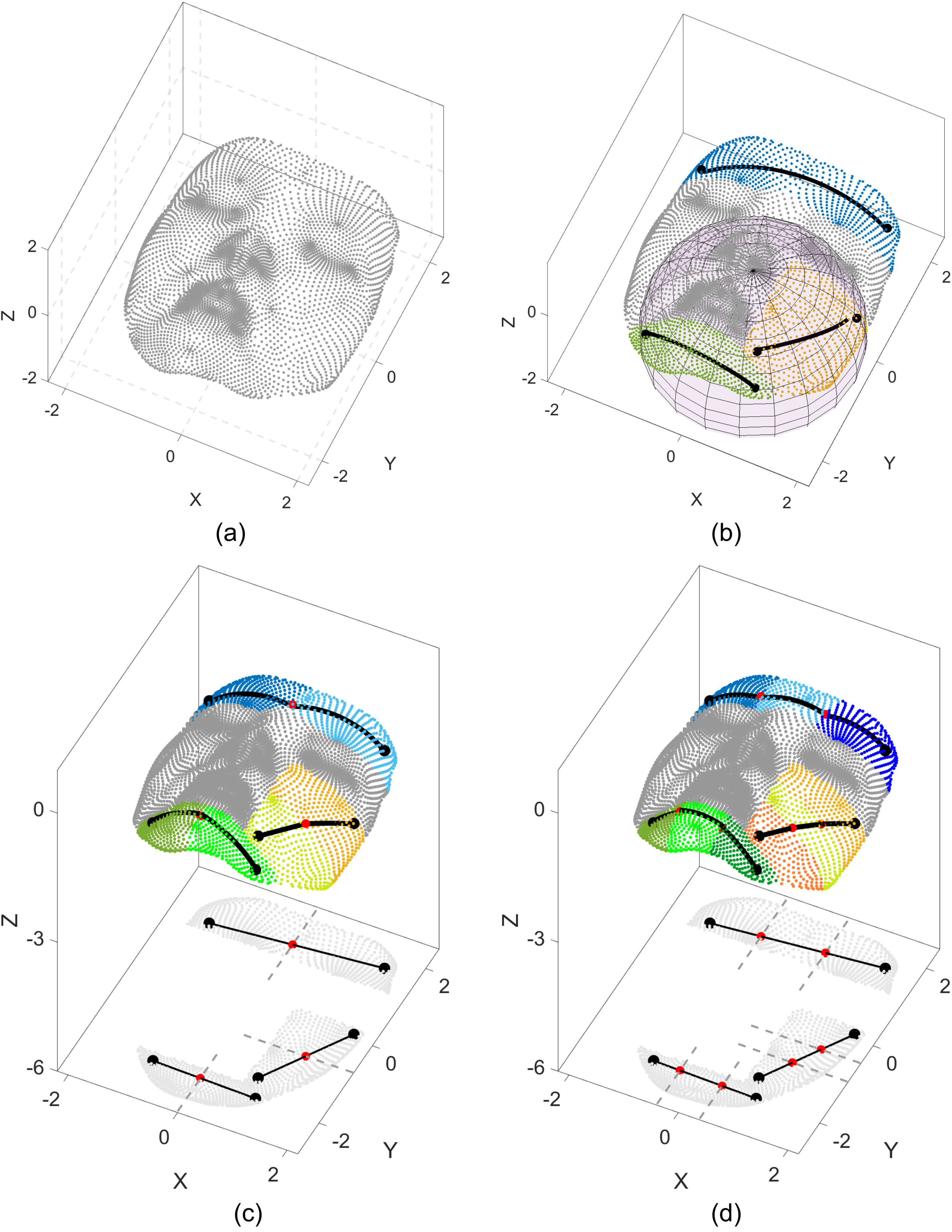}
    \caption{Illustration of geodesic distance calculation using the spherelet approach on the forehead, cheek, and chin of Nefertiti's face. 
    (a) The plot of point cloud data of Nefertiti's face on which we calculate the geodesic distances with 
    (b) one spherelet, where we denote the terminal points with black dots and exemplify the osculating sphere for the cheek in purple, 
    (c) two spherelets, where we denote the segmentation and intersection points with red dots and plot grid lines with dashed grey lines, and 
    (d) three spherelets.}
    \label{fig:Nefertiti}
\end{figure}
\begin{table}[t]
\small
\aboverulesep=0ex 
\belowrulesep=0ex 
\renewcommand{\arraystretch}{1.2} 
\setlength{\tabcolsep}{3.pt}
\caption{Geodesic Distance Calculation Results.}
\centering
\begin{tabular}{cc|c|c|c|c|c}
\toprule
\multicolumn{1}{c}{} & \multicolumn{2}{c}{\textbf{Forehead}} & \multicolumn{2}{c}{\textbf{Cheek}} & \multicolumn{2}{c}{\textbf{Chin}} \\
\cmidrule(rl){2-3} \cmidrule(rl){4-5} \cmidrule(rl){6-7}
\textbf{Method} & \multicolumn{1}{c|}{$\SI{}{\mm}$} & \multicolumn{1}{c}{\si{\second}}& \multicolumn{1}{c|}{$\SI{}{\mm}$} & \multicolumn{1}{c}{\si{\second}}& \multicolumn{1}{c|}{$\SI{}{\mm}$} & \multicolumn{1}{c}{\si{\second}}   \\
\hline
IsoMap     & $3.386$ & $0.820$   & $2.108$  & $0.389$   & $2.562$  & $0.373$ \\
MDS        & $3.3746$   & $0.101$   & $2.0489$   & $0.077$   & $2.4433$    & $0.071$ \\
Heat       & $3.448$    & $0.699$   & $2.1831$   & $0.626$   & $2.564$     & $0.599$ \\
Eikonal    & $3.3748$   & $0.603$   & $2.0583$   & $0.501$   & $2.4433$    & $0.459$\\
\textbf{OneSph. (ours)}   & $3.2992$   & $0.018$   & $1.9616$   & $0.017$   & $2.4876$    & $0.018$ \\
\textbf{TwoSph. (ours)}   & $3.3128$   & $0.026$   & $2.0070$   & $0.023$   & $2.5196$    & $0.023$ \\
\textbf{ThreeSph. (ours)} & $3.3308$   & $0.025$   & $2.0223$   & $0.023$   & $2.5370$    & $0.027$\\
\hline          
\end{tabular}
\label{Tab_GeoDisTest}
\end{table}
\subsection{Geodesic Distance Calculation}\label{sec_geodist}
In this section, we showcase the efficacy of the spherelets-based approach for calculating geodesic distance, using the face of \textit{Nefertiti} as an illustrative example (refer to Fig.~\ref{fig:Nefertiti}(a)).
Our goal is to calculate the geodesic distance on Nefertiti's face surface that is expressed by the corresponding point cloud data.
To begin with, we align the facial data points to ensure that the least principal component aligns parallel to the $z$-axis, and the face surface is symmetrically positioned with respect to the $y$-axis.
For the projection plane, we select $z=-6$, which is parallel to the $x$-$y$ plane.
In each region of the forehead, cheek, and chin, we approximate the surface patch with one spherelet, two spherelets, and three spherelets, respectively.
The geodesic distance calculated in each region using different numbers of spherelets is shown in Fig.~\ref{fig:Nefertiti}(b)-(d).

\begin{figure}[t]
    \centering
    \includegraphics[width=0.95\columnwidth]{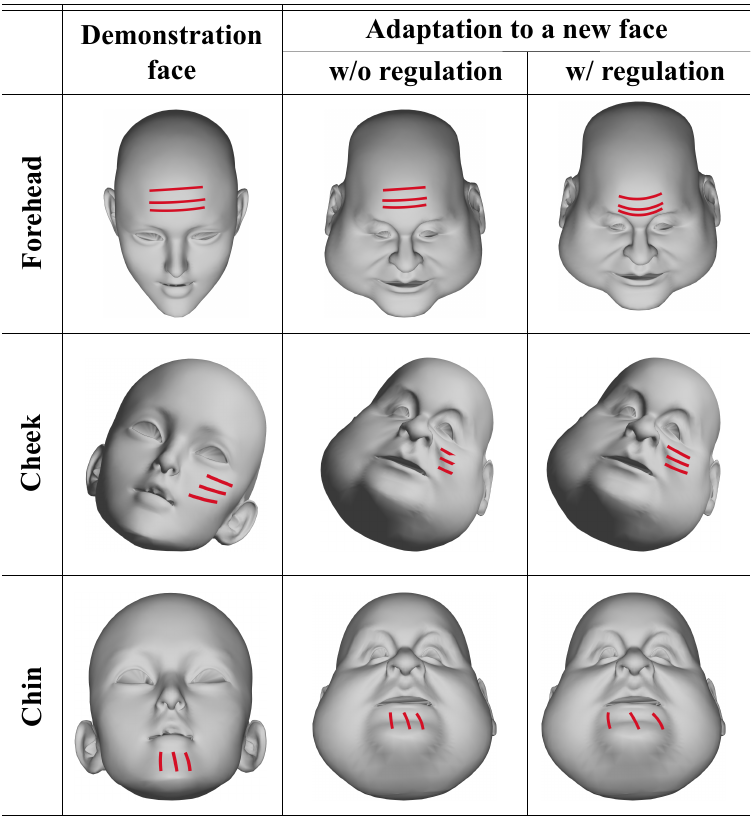}
    \caption{Illustration of reference trajectory on the demonstration face (\textit{left column}), its reproduction on a new face without regulation (\textit{middle column}), and its reproduction on a new face with (\textit{right column}).}
    \label{fig:EXP_prob_reg}
\end{figure}

For comparison, we also calculate the geodesic distances between the same endpoints using other methods, including ISOMAP (with the Floyd algorithm for pairwise geodesic distances), MDS, Heat Flow, and the Eikonal approach. 
Besides comparing the geodesic distances, we also record the total required running time.
Table~\ref{Tab_GeoDisTest} summarizes the comparison results.
It can be seen that our approach achieves on-par performance in the meanwhile possessing the merit of considerably reducing the computational time.

\subsection{Skill Adaptation to New Faces}\label{sec_skilladapt}
Here we evaluate the effectiveness of the probabilistic non-rigid registration technique.
The goal is to justify the necessity of imposing additional regulation effects when performing an adaptation of a trajectory from the demonstration face to a new face.
We investigate the adaptation behaviors on the regions of the forehead, cheek, and chin whose shapes are determined using nine, seven, and six landmarks, respectively.

An illustration of the regulation effects is depicted in Fig.~\ref{fig:EXP_prob_reg}.
During the evaluations, we set the smoothing parameter to be $\lambda = 0.5$.
For trajectory modulation on the forehead, the control point besides the hairline is assigned with a weight matrix of $20\mathbf{I}$, thus prohibiting the adapted trajectory from the potential interference with the subject's hair.
For adaptation on the cheek region, we improve the occlusion between the adapted trajectories and the cheek surface by setting the cheek tip control point with a weight of $1\mathrm{e}2\mathbf{I}$.
Regarding the chin region, we choose the weight matrix as $10\mathbf{I}$ for the critical control points, stretching the demonstration trajectories to fit the wide chin of the new face. 

\begin{figure}[t]
    \centering
    \includegraphics[width=\columnwidth]{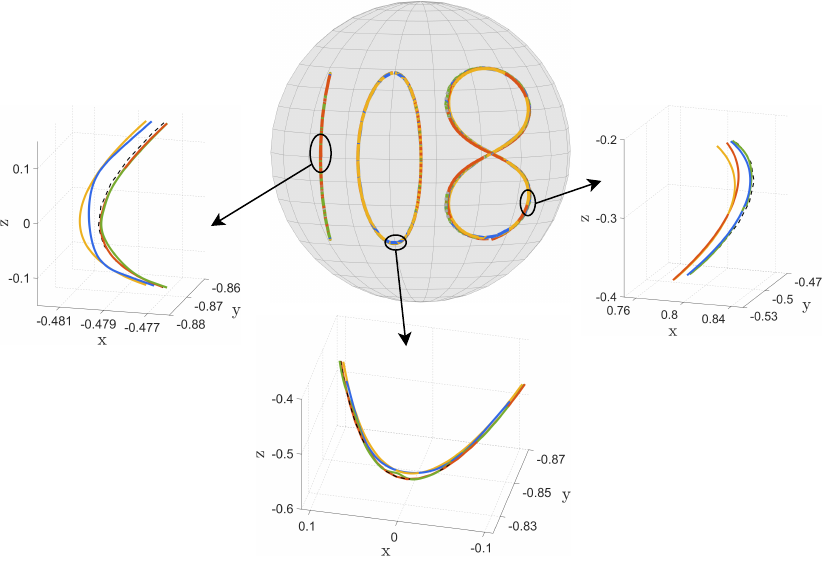}
    \caption{Schematic illustration of the learning results with different imitation algorithms. The demonstrated $'1'$, $'0'$, and $'8'$-shaped periodic trajectories (dashed black line) are separately learned by our approach (green), GA-DMP(yellow), R-LQT (red), and Ori-ProMP (blue), respectively.}
    \label{fig:benchmark}
\end{figure}

\begin{table}[t]
\small
\aboverulesep=0ex 
\belowrulesep=0ex 
\renewcommand{\arraystretch}{1.2}
\setlength{\tabcolsep}{9.9pt} 
\caption{Learning Errors of Different Motion Imitation Algorithms.}
\centering
\begin{tabular}{cccc}
\toprule
& $"1"$ & $"0"$ &  $"8"$ \\
\hline
GA-DMP~\cite{abu2020geometry}     & $0.0371$  & $0.0406 $  & $0.0572$ \\
R-LQT~\cite{calinon2020gaussians} & $0.0144$  & $0.0224$   & $0.0312$ \\
Ori-ProMP~\cite{rozo2022orientation}    & $0.0322$  & $0.0161$   & $0.0124$ \\
\textbf{Our approach}                      & $\mathbf{0.0065}$  & $\mathbf{0.0075}$   & $\mathbf{0.0022}$ \\
\hline
\end{tabular}
\label{Tab_CompBaselines}
\end{table}

\subsection{Comparison with Baselines}\label{sec_compsota}
In this section, we compare our proposed approach with several representative state-of-the-art movement primitive algorithms in terms of imitating geometric motion with periodicity.
Specifically, for benchmark baselines, we select Geometry-Aware Dynamics Movement Primitives (GA-DMP)~\cite{abu2020geometry}, Riemannian Linear Quadratic Tracking (R-LQT)~\cite{calinon2020gaussians}, and Orientation Probabilistic Movement Primitives (Ori-ProMP)~\cite{rozo2022orientation} that represent the geometry-aware counterparts of the classical dynamics movement primitives, linear quadratic tracking, and probabilistic movement primitives. 
We specify three periodic demonstration trajectories that evolve on a unit sphere to imitate, namely $'1'$, $'0'$, and $'8'$.

For our approach, we use the PER kernel, and the relevant parameters are selected to be $\sigma_p = 1$, $l_p = 2$, and $p = 100$.
We use $30$ Von-Mises basis functions per dimension for GA-DMP, which is defined as
$\exp({\cos(2\pi(t-c_i))}/{h})$,
where $h = 5$ defines the width of the basis function and $c_i$ determines the uniformly distributed center of the $i$-th basis function within the period range. 
For imitating with R-LQT, we set the penalization matrix on the control inputs to be $0.2\mathbf{I}$ and the time window is $30$ steps. 
Regarding Ori-ProMP, the number of basis functions is $30$ and the learning rate is $0.05$.

The learning performances by all the algorithms are depicted in Fig.~\ref{fig:benchmark}, which implies that all the algorithms can truthfully reproduce the demonstrated trajectories.
We then quantify the imitation performances based on the imitation metric of \eqref{eq_evalreprometric}, and the obtained numerical results are summarized in Table~\ref{Tab_CompBaselines}, revealing that our approach exhibits a high level of imitation fidelity.

\begin{figure}[t]
    \centering
    \includegraphics[width=0.9\columnwidth]{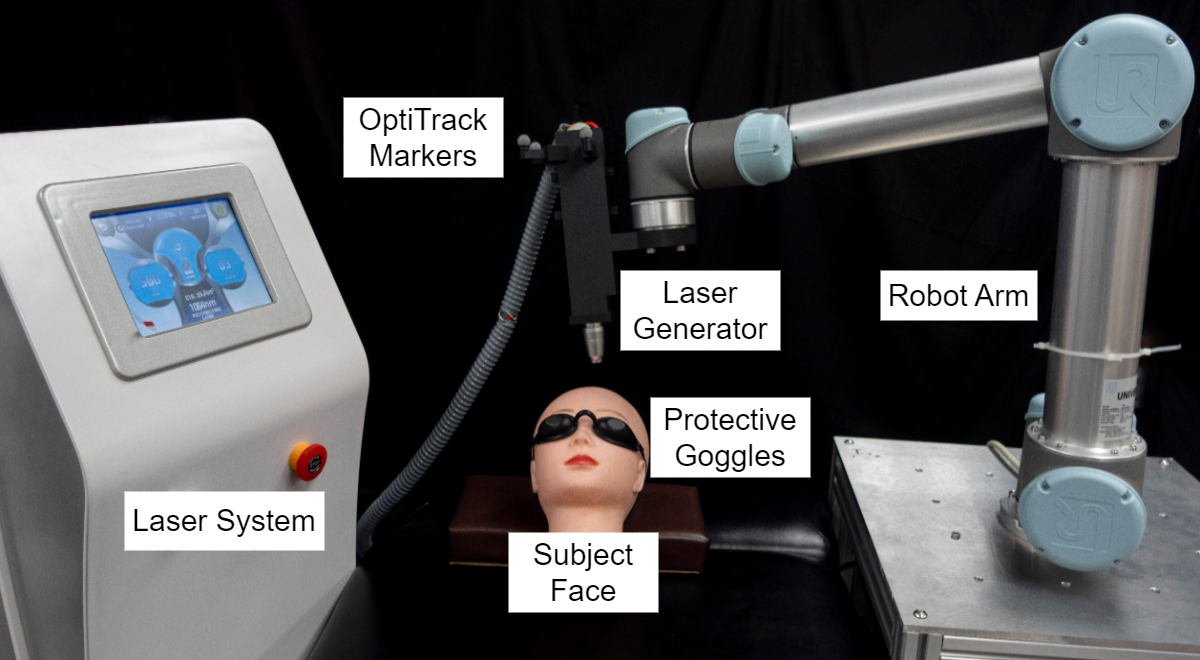}
    \caption{Illustration of the experimental setup for performing robotic cosmetic dermatology using photorejuvenation.}
    \label{fig:expsetup}
\end{figure}

\begin{figure}[t]
    \centering
    \includegraphics[width=0.99\columnwidth]{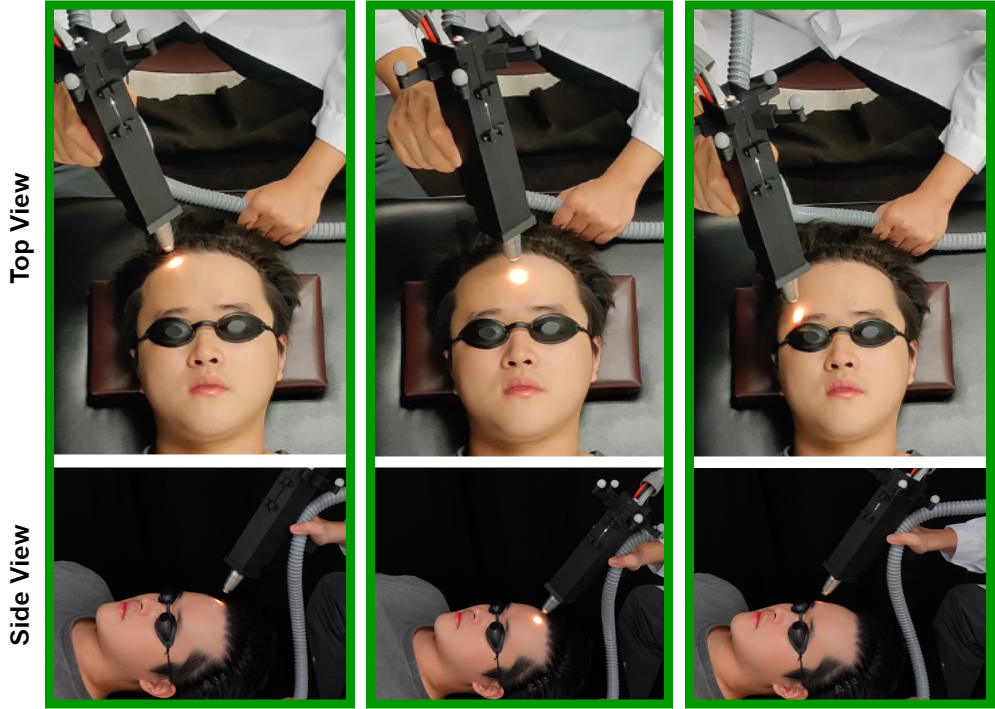}
    \caption{Snapshots of the demonstration procedure where the treatment motion on a human forehead is transferred by passive observation.}
    \label{fig:humandemo}
\end{figure}

\subsection{Skin Photorejuvenation Experiments}\label{sec_realexp}
In this section, we conduct real experiments to learn the treatment trajectory of skin photo-rejuvenation from demonstrations.
Fig.~\ref{fig:expsetup} presents the overall experimental setup.
A UR5 robot arm from Universal Robots is employed, and a laser cosmetic instrument fixed with a 3D-printed fixation case is attached to its end-effector.
For the sensing system, we use an Intel RealSense D405 RGBD camera to scan the subject's face for the point cloud data.
Also, the OptiTrack motion capture system is used to record the demonstrated trajectory with the markers attached to the laser cosmetic instrument.
For the laser system, the pulse frequency of the laser shots is controlled with an on-board relay.
We set constant values for the pulse frequency with $\SI{10}{\hertz}$.
In addtion, we set the thermal power as $\SI{100}{\milli\joule}$, and spot diameter as $\SI{10}{\milli\meter}$. 
Throughout the experiment, the involved human subjects\footnote{Ethics Approval Reference Number: HSEARS20201202001, Human Subjects Ethics Sub-committee, Departmental Research Committee, The Hong
Kong Polytechnics University, Hong Kong.} exposed to the laser pulses are asked to wear a pair of protective goggles.

During the skill transfer phase, the point cloud data of the subject's face is first obtained by the depth camera. 
Subsequently, a dermatologist personally demonstrates the desired treatment trajectory by manipulating the laser cosmetic instrument, which is recorded by the motion capture cameras.
During the treatment, the involved subject is required to remain in a still supine position.
The demonstration procedure by passive observation can be seen in Fig.~\ref{fig:humandemo}.
\begin{figure}[t]
    \centering
    \includegraphics[width=0.99\columnwidth]{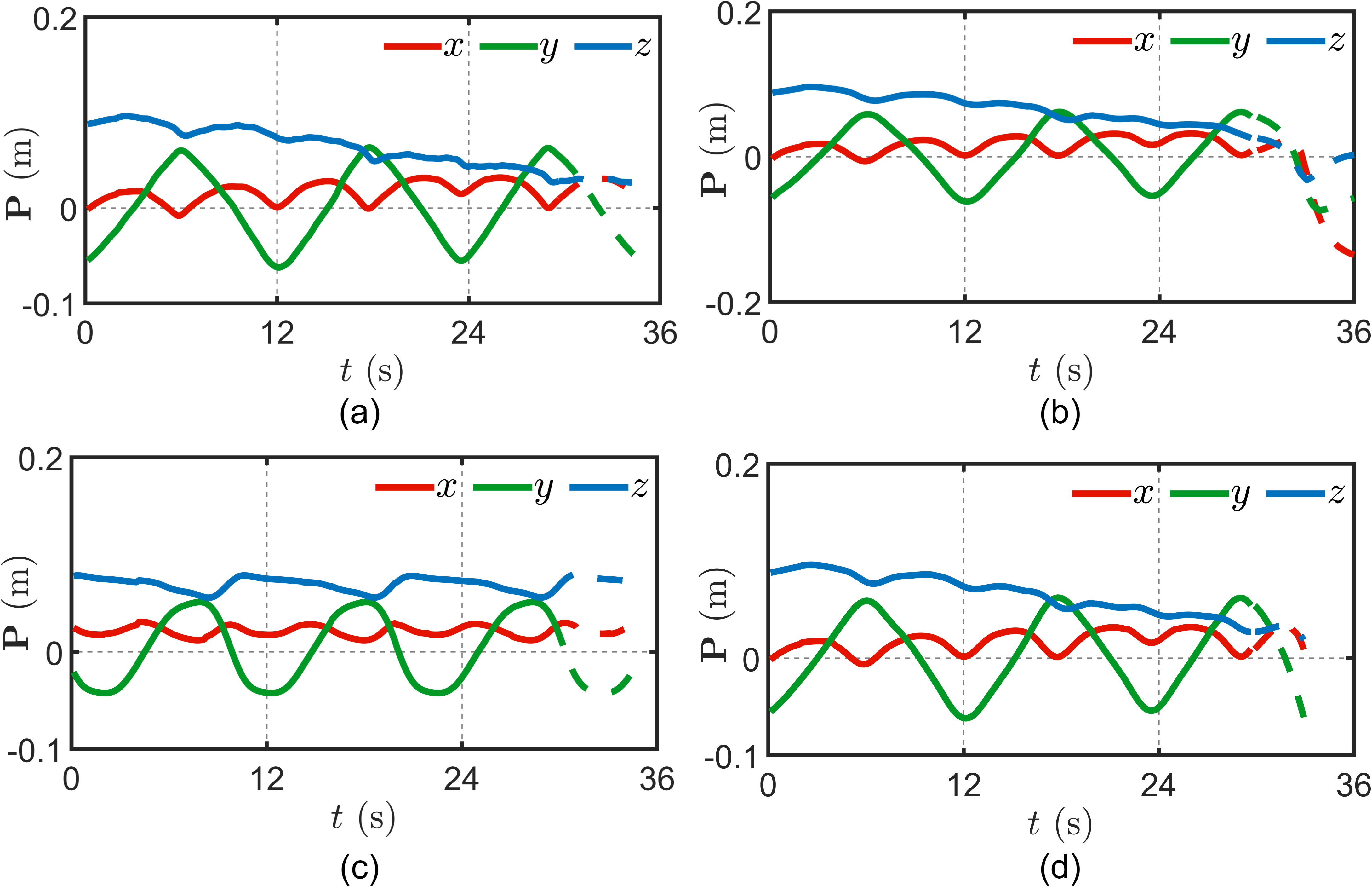}
    \caption{Illustration of temporal evolutions of the trajectory of (a) demonstration, and reproduction with the (b) SE kernel (c) PER kernel, and (d) QP kernel where solid lines denote the reproduction trajectory and dashed lines denote the generalization trajectory.}
    \label{fig:exptemptraj}
\end{figure}

\begin{figure}[t]
    \centering
    \includegraphics[width=0.99\columnwidth]{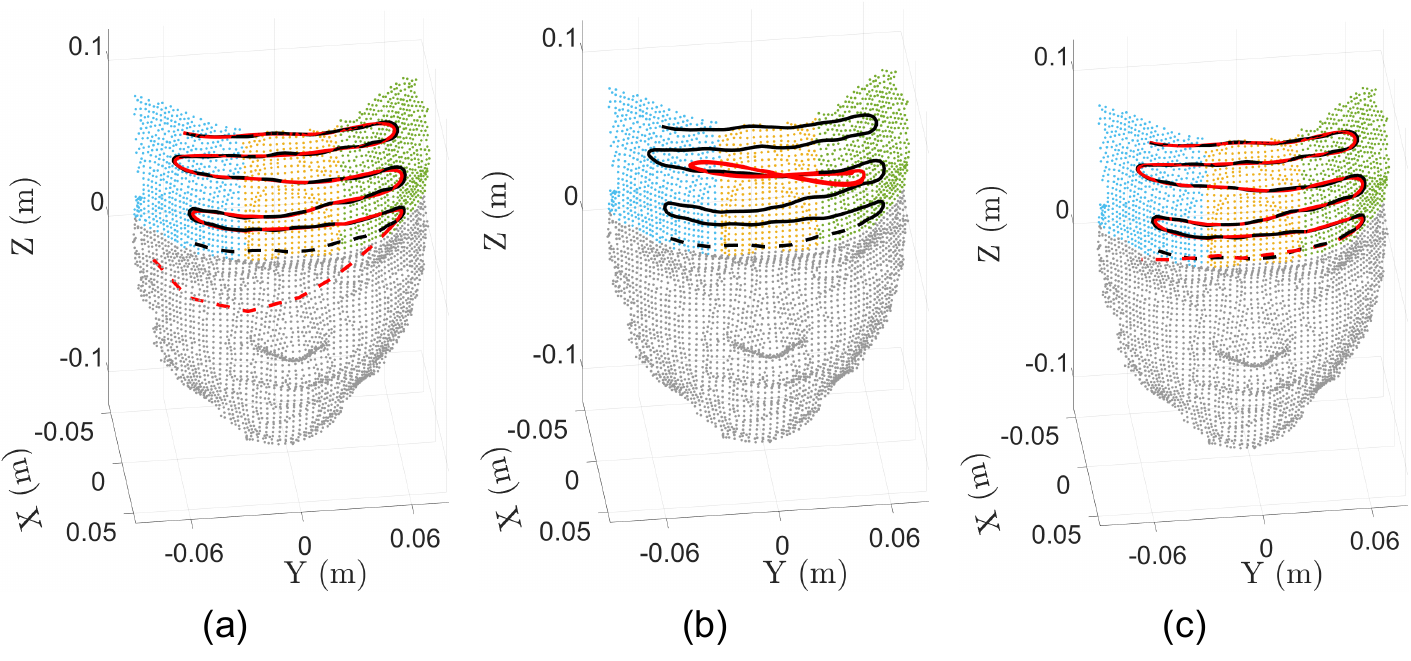}
    \caption{Illustration of the learned geometric trajectory (red) from the demonstrated motion (black) by (a) SE kernel (b) PER kernel, and (c) QP kernel where solid lines denote the reproduction trajectory and dashed lines denote the generalization trajectory, and the segmentation with spherelets is represented by different colors.}
    \label{fig:reproker}
\end{figure}

\begin{figure}[t]
    \centering
    \includegraphics[width=0.99\columnwidth]{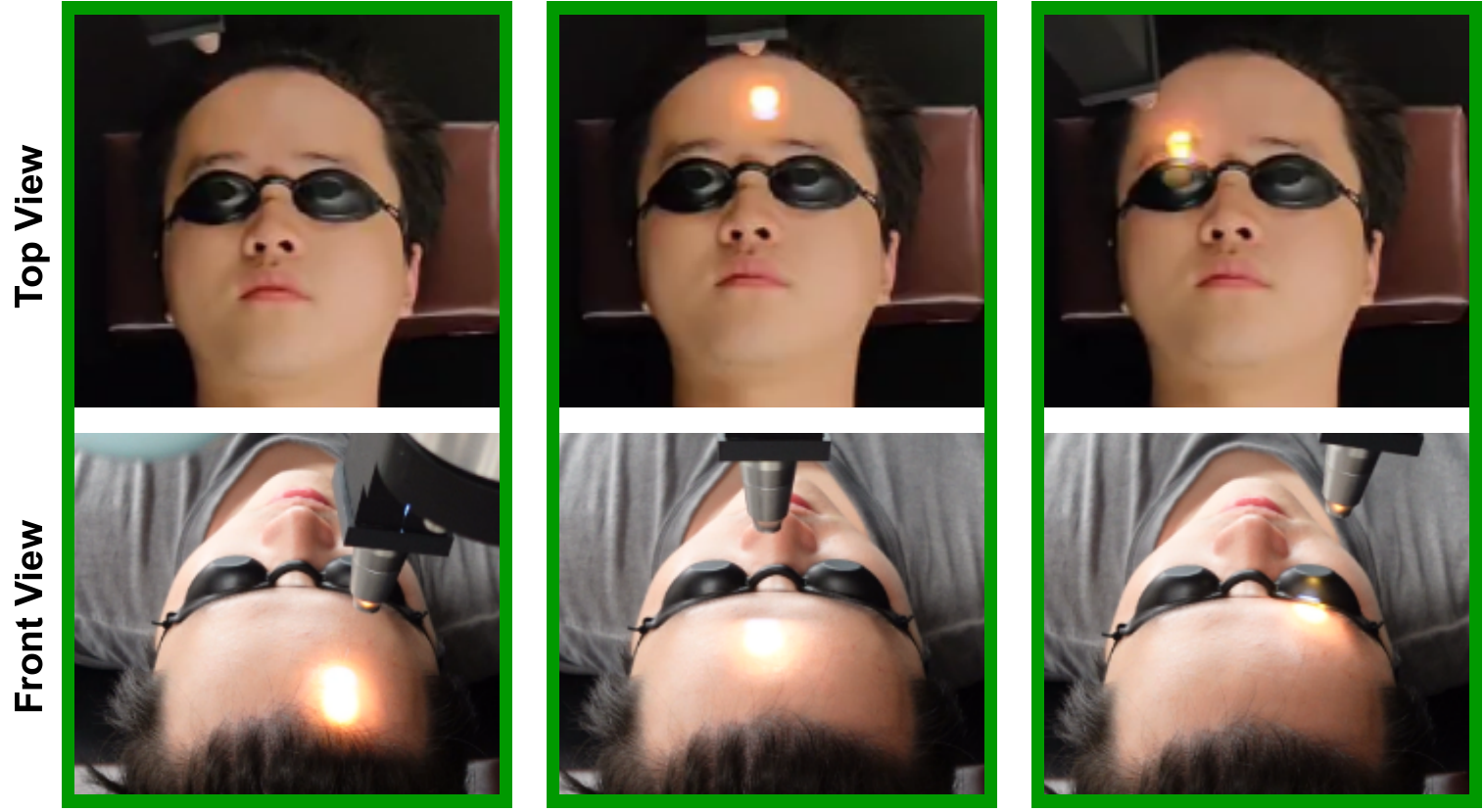}
    \caption{Snapshots of reproduction of the treatment skill on the demonstration face with a robot arm.}
    \label{fig:exprepro}
\end{figure}

\begin{table}[t]
\small
\aboverulesep=0ex 
\belowrulesep=0ex 
\renewcommand{\arraystretch}{1.2}
\setlength{\tabcolsep}{9.9pt} 
\caption{Learning Errors with Different Kernels.}
\centering
\begin{tabular}{cccc}
\toprule
& SE & PER &  QP \\
\hline
Reproduction     & $0.001$  & $0.03$  & $\mathbf{0.0008}$ \\
Generalization   & $0.125$  & $0.06$  & $\mathbf{0.01}$ \\
\hline
\end{tabular}
\label{Exp_Compkernels}
\end{table}

\begin{figure}[t]
    \centering
    \includegraphics[width=0.99\columnwidth]{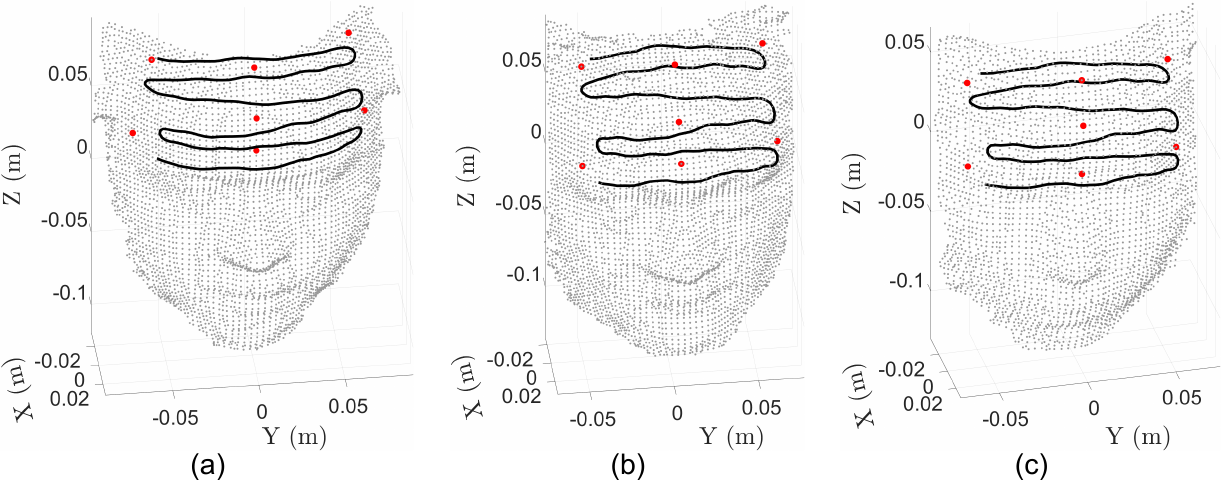}
    \caption{Illustration of adapting the laser trajectory on the demonstration face (a) towards other unseen faces (b) and (c) where red dots denote the landmarks used for performing trajectory transfer.}
    \label{fig:adaptfaces}
\end{figure}

We first evaluate the reproduction performance in terms of treating the same subject as in the demonstration. 
Likewise, we compare the learning performance using the SE, PER, and QP kernels to learn the demonstrated motion pattern.
The learning results are shown in Fig.~\ref{fig:exptemptraj}.
Besides, the trajectory evolution on the facial point cloud is shown in Fig.~\ref{fig:reproker}.
It can be seen that the trajectory prediction by the QP kernel attains the best performance as the demonstration trajectory exhibits a quasi-periodic motion pattern.
The quantitative evidence of the learning performances by all three kernels is summarized in Table~\ref{Exp_Compkernels}, where the evaluations for motion reproduction and generalization are conducted regarding \eqref{eq_evalreprometric} and \eqref{eq_evalgenmetric}, respectively.
Once the trajectory is learned with the QP kernel, we then reproduce the treatment trajectory on the subject, as shown in Fig.~\ref{fig:exprepro}.

\begin{figure}[t]
    \centering
    \includegraphics[width=0.99\columnwidth]{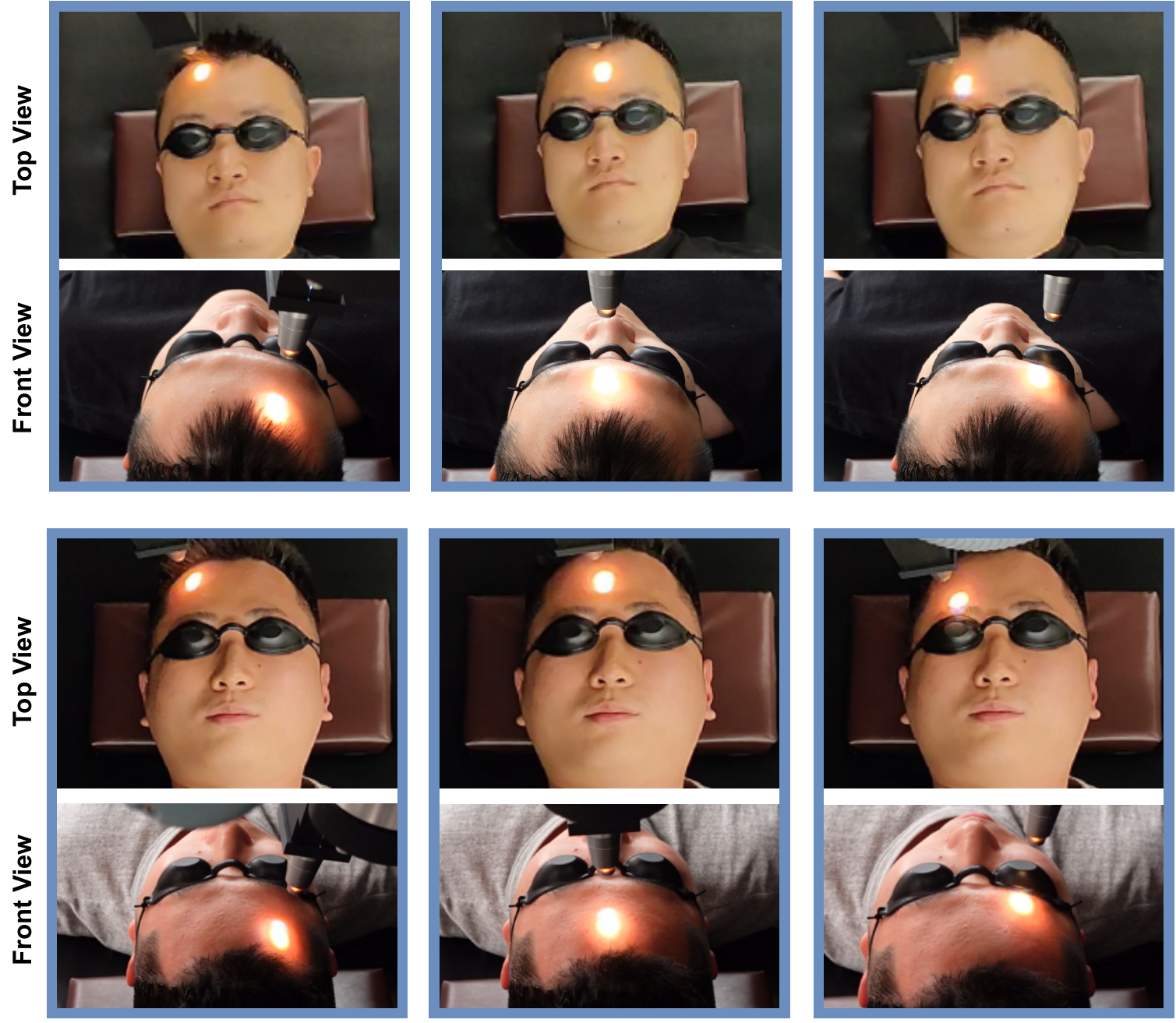}
    \caption{Snapshots of adaptation of the demonstrated treatment skill to other new faces that are not involved in the demonstration.}
    \label{fig:expadapt}
\end{figure}

Finally, we study the issue of adapting the learned trajectory from the face in the demonstration to the faces that were unseen before.
When a new subject comes to receive the treatment, we first scan the face for the corresponding point cloud data. 
Afterwards, we select the landmarks on the point cloud that are used for transferring the learned demonstration trajectory.
Fig.~\ref{fig:adaptfaces} shows the point cloud for the demonstration face and two new subjects' faces.
The red dots mark the source control points on the demonstration face as well as the target control points on the new faces.
The procedure of the adaptation treatment is shown in Fig.~\ref{fig:expadapt}.

\section{Conclusion}\label{conclusion}
In this paper, we have addressed the issue of robotic cosmetic dermatology using the learning-by-demonstration paradigm. 
Specifically, we have tackled the challenge of motion imitation through the lens of structured prediction, which is powerful in handling geometry-structured data arising from facial surfaces. 
Additionally, we have developed an adaptation strategy based on the non-registration technique to treat the facial surface of a new subject, which was not observed during the demonstration.
The real-world experiments have shown our proposed method's effectiveness in performing photorejuvenation for cosmetic dermatology.

There are still some limitations associated with our proposed approach. 
For instance, in the current setup, we require the human subject to maintain a static position throughout the treatment. 
Ideally, it would be favorable to incorporate reactive behaviors in the robot to respond to potential movements from the region of interest~\cite{zhou2023neural}.
Undoubtedly, the integration of robotic technologies into the beauty industry holds promise, and we plan to continue our efforts in this direction.
For future work, we intend to conduct further tests to obtain quantitative evidence regarding the administration of thermal doses for enhancing skin conditions; We are currently working on the development of new thermal servoing controls (e.g., as in \cite{luyin_tro2022}) for these types of laser-based procedures.

{\appendices
\section{Notation}\label{Notation}
The notation used throughout the paper is in Table~\ref{tab:notation}.

\begin{table}[t]
\centering
\caption{Summary of Key Notations.}
\label{tab:notation} 
\begin{tabular}{c || c c}
  \hline
  & Notation & Description \\
  \hline
  \multirow{17}{*}{\rotatebox[origin=c]{90}{Math}} & $(\cdot)^\intercal$ &  Transpose operator \\
   & $\otimes$ &  Kronecker product \\
   & $\odot$ & Hadamard product\\
   & $\mathcal{H}$ & Hilbert space \\
   & $\circ$ & Function composition \\
   & $\mathbf{I}$ &  The identity matrix \\
   & $\langle \cdot, \cdot \rangle_\mathcal{H}$ & Inner product of $\mathcal{H}$ \\
   & $\boldsymbol{\Gamma}$  & Parallel transport\\
   & $\mathtt{Exp}$    & Exponential mapping\\
   & $\nabla_{\mathcal{M}}$, $\nabla$ & Riemannian, Euclidean gradient\\
   & $\| \cdot \|_{\mathtt{Frob}}$ & Frobenius norm\\
   & $\mathtt{Tr}(\cdot)$  &  Trace of a matrix \\
   & $\mathtt{blockdiag}(\cdot)$ &   Block-diagonal concatenation\\ 
   & $\mathtt{Proj}$ & Orthogonal projector\\  
   & $\mathbf{Q}, \mathbf{R}$ & QR composition matrices \\
   \hline
  \multirow{14}{*}{\rotatebox[origin=c]{90}{Learning}} 
   & $\mathbb{T}$ & Dataset of demonstration  \\
   & $\mathbf{c}$, $\mathbf{c}^{-1}$  & Decoding, encoding rule\\
   & $\mathbf{g}$ & Surrogate mapping function \\
   & $\mathbf{s}$ & Predicted structured output \\
   & $\mathcal{L}$ &  Surrogate loss \\
   & $\Delta$ & SELF loss function \\
   & $\mathbf{V}$  & Continuous linear operator\\
   & $\boldsymbol{\Psi}$ &  Feature map \\
   & $k(\cdot, \cdot)$ & Kernel function \\
   & $\sigma, l$ &  Hyperparameters of kernel \\
   & $\bm{\mathcal{K}}$ & Matrix-valued kernel \\
   & $\mathbf{K}$ & Gram matrix\\
   & $\boldsymbol{\alpha}$ & Score function \\

   \hline
   \multirow{15}{*}{\rotatebox[origin=c]{90}{Robot}} 
   & $\mathcal{S}$ & Surface manifold\\
   & $\mathbb{S}^2$ & Sphere manifold \\
   &  $t$ & Trajectory time stamp \\
   &  $\mathbf{p}$ & Cartesian position \\ 
   &  $T$    & Trajectory period\\
   & $\mathbf{O}$, $r$ & Center, radius of osculating sphere\\
   & $C(\mathbf{O}, r)$ & Algebraic fitting loss\\
   & $\mathbb{F}$ &  Point cloud of facial surface \\
   & $\boldsymbol{l}^p_{v}$, $\boldsymbol{l}^p_{h}$ & Vertical, horizontal grid lines \\
   & $\boldsymbol{\xi}^s$, $\boldsymbol{\xi}^p$ & Segmentation, intersection points \\
   & $\boldsymbol{\Pi}_p^s(\cdot)$ & Mapping from projection plane to sphere \\
   & $\boldsymbol{f}(\cdot)$ & Adaptation rule\\ 
   & $E(\cdot)$ & Bending energy\\
   & $\hat{\boldsymbol{\xi}}$, $\hat{\boldsymbol{\xi}}_i'$ & Source, target control points\\
   & $\boldsymbol{\rho}$     & Thin-plate spline basis\\
  \hline
\end{tabular}
\end{table}

\section{Proof of Proposition \ref{Quasi_periodic}}\label{app_quasi}
In this section, we show that the behavior of the trajectories in the surrogate space can also lead to the corresponding behavior in the structured space that has a loss function of the geodesic distance.     
We first illustrate the arithmetic case.
For convenience, we denote
\begin{equation}
a+1 := a^+, \;  a-1 := a^-, \; \text{and} \;\; \mathbf{P}(\tau+aT) := \mathbf{P}_{a}.
\end{equation}

From~\eqref{geoarith}, we equivalently have the following
\begin{subequations}
\begin{align}
\Delta(\mathbf{P}_{a^-}, \mathbf{P}_{a}) &= \Delta(\mathbf{P}_{a}, \mathbf{P}_{a^+}) = \|\mathfrak{C}(\tau)\|, \label{eqdist}\\
\Delta(\mathbf{P}_{a^-}, \mathbf{P}_{a^+}) &= \Delta(\mathbf{P}_{a^-}, \mathbf{P}_{a}) + \Delta(\mathbf{P}_{a}, \mathbf{P}_{a^+}). \label{cogeoln}
\end{align}
\end{subequations}

By using~\eqref{defSELF}, we can express~\eqref{eqdist} as
\begin{equation}\label{eqbySelf}
\langle \boldsymbol{\Psi}(\mathbf{P}_{a^-}), \mathbf{V}\boldsymbol{\Psi}(\mathbf{P}_{a}) \rangle_{\mathcal{H}} 
= \langle \boldsymbol{\Psi}(\mathbf{P}_{a}), \mathbf{V}\boldsymbol{\Psi}(\mathbf{P}_{a^+}) \rangle_{\mathcal{H}}.   
\end{equation}
Denoting $\delta \boldsymbol{\Psi}_a := \boldsymbol{\Psi}(\mathbf{P}_{a})-\boldsymbol{\Psi}(\mathbf{P}_{a^-})$, the LHS of~\eqref{eqbySelf} can be shown to be
\begin{subequations}
\begin{align}
&\langle \boldsymbol{\Psi}(\mathbf{P}_{a^-}), \mathbf{V}\boldsymbol{\Psi}(\mathbf{P}_{a}) \rangle_{\mathcal{H}} \\
=& \langle \boldsymbol{\Psi}(\mathbf{P}_{a^-}), \mathbf{V}(\boldsymbol{\Psi}(\mathbf{P}_{a^-})+\delta \boldsymbol{\Psi}_a) \rangle_{\mathcal{H}}\\
=& \langle \boldsymbol{\Psi}(\mathbf{P}_{a^-}), \mathbf{V}\delta \boldsymbol{\Psi}_a \rangle_{\mathcal{H}}\\
=& \langle \boldsymbol{\Psi}(\mathbf{P}_{a})-\delta\boldsymbol{\Psi}_a, \mathbf{V}\delta\boldsymbol{\Psi}_a \rangle_{\mathcal{H}}\\
=& \langle \boldsymbol{\Psi}(\mathbf{P}_{a}), \mathbf{V}\delta\boldsymbol{\Psi}_a \rangle_{\mathcal{H}},\label{LHSari}
\end{align}  
\end{subequations}     
where we have used the linearity of the inner product and the fact that $\langle \boldsymbol{\Psi}(\mathbf{p}), \mathbf{V}\boldsymbol{\Psi}(\mathbf{p})\rangle_\mathcal{H}=\Delta(\mathbf{p}, \mathbf{p}) = 0$. 
Also, the RHS of~\eqref{eqbySelf} can be shown to be
\begin{subequations}
\begin{align}
&\langle \boldsymbol{\Psi}(\mathbf{P}_{a}), \mathbf{V}\boldsymbol{\Psi}(\mathbf{P}_{a^+}) \rangle_{\mathcal{H}}\\
=& \langle \boldsymbol{\Psi}(\mathbf{P}_{a}), \mathbf{V}(\boldsymbol{\Psi}(\mathbf{P}_{a})+\delta\boldsymbol{\Psi}_{a^+}) \rangle_{\mathcal{H}}\\
=& \langle \boldsymbol{\Psi}(\mathbf{P}_{a}), \mathbf{V}\delta\boldsymbol{\Psi}_{a^+} \rangle_{\mathcal{H}}.\label{RHSari}
\end{align}  
\end{subequations}  
From~\eqref{eqbySelf}, we then equate \eqref{LHSari} and \eqref{RHSari}, leading to 
\begin{equation}\label{arigeodisteq}
\langle \boldsymbol{\Psi}(\mathbf{P}_{a}), \mathbf{V}\delta\boldsymbol{\Psi}_a \rangle_{\mathcal{H}}    = \langle \boldsymbol{\Psi}(\mathbf{P}_{a}), \mathbf{V}\delta\boldsymbol{\Psi}_{a^+} \rangle_{\mathcal{H}}.
\end{equation}

For~\eqref{cogeoln}, its LHS can be written as
\begin{subequations}
\begin{align}
&\langle \boldsymbol{\Psi}(\mathbf{P}_{a^-}), \mathbf{V}\boldsymbol{\Psi}(\mathbf{P}_{a^+}) \rangle_{\mathcal{H}}\\ 
=& \langle \boldsymbol{\Psi}(\mathbf{P}_{a})-\delta\boldsymbol{\Psi}_a, \mathbf{V}(\boldsymbol{\Psi}(\mathbf{P}_{a})+\delta\boldsymbol{\Psi}_{a^+}) \rangle_{\mathcal{H}}\\
=& \langle \boldsymbol{\Psi}(\mathbf{P}_{a}), \mathbf{V}\delta\boldsymbol{\Psi}_{a^+} \rangle_{\mathcal{H}} - 
\langle \delta\boldsymbol{\Psi}_a, \mathbf{V}\boldsymbol{\Psi}(\mathbf{P}_{a}) \rangle_{\mathcal{H}} - \notag\\
&\qquad\qquad\qquad\qquad\qquad\qquad\langle \delta\boldsymbol{\Psi}_a, \mathbf{V}\delta\boldsymbol{\Psi}_{a^+} \rangle_{\mathcal{H}}.  \label{geodistLHSari}
\end{align}
\end{subequations}  
Also, the RHS of~\eqref{cogeoln} can be written as
\begin{subequations}
\begin{align}
&\langle \boldsymbol{\Psi}(\mathbf{P}_{a^-}), \mathbf{V}\boldsymbol{\Psi}(\mathbf{P}_{a}) \rangle_{\mathcal{H}} + \langle \boldsymbol{\Psi}(\mathbf{P}_{a}), \mathbf{V}\boldsymbol{\Psi}(\mathbf{P}_{a^+}) \rangle_{\mathcal{H}}\\
=& \langle \boldsymbol{\Psi}(\mathbf{P}_{a})-\delta\boldsymbol{\Psi}_a, \mathbf{V}\boldsymbol{\Psi}(\mathbf{P}_{a}) \rangle_{\mathcal{H}} + \langle \boldsymbol{\Psi}(\mathbf{P}_{a}), \mathbf{V}\delta\boldsymbol{\Psi}_{a^+}\rangle_{\mathcal{H}}\\
=& -\langle \delta\boldsymbol{\Psi}_a, \mathbf{V}\boldsymbol{\Psi}(\mathbf{P}_{a}) \rangle_{\mathcal{H}} + \langle \boldsymbol{\Psi}(\mathbf{P}_{a}), \mathbf{V}\delta\boldsymbol{\Psi}_{a^+} \rangle_{\mathcal{H}}. \label{geodistRHSari}
\end{align}
\end{subequations} 

By equating \eqref{geodistLHSari} and \eqref{geodistRHSari} due to \eqref{cogeoln}, we then have
\begin{equation}\label{aricogeodist}
\langle \delta\boldsymbol{\Psi}_a, \mathbf{V}\delta\boldsymbol{\Psi}_{a^+} \rangle_{\mathcal{H}} = 0.   
\end{equation}

It can be seen that the condition $\delta\boldsymbol{\Psi}_{a} = \delta\boldsymbol{\Psi}_{a^+} := \delta\boldsymbol{\Psi}_{\tau}$ is sufficient to make both \eqref{arigeodisteq} and \eqref{aricogeodist}
hold, implying that the following arithmetic pattern in the surrogate space leads to~\eqref{geoarith}. 
\begin{equation}
\boldsymbol{\Psi}(\mathbf{P}_{a})  = \boldsymbol{\Psi}(\mathbf{P}_{a^-}) +\delta \boldsymbol{\Psi}_{\tau}.      
\end{equation}

For the cumulative case~\eqref{geocumu}, we equivalently have
\begin{subequations}
\begin{align}
\Delta(\mathbf{P}_{a^-}, \mathbf{P}_{a})/a &= \Delta(\mathbf{P}_{a}, \mathbf{P}_{a^+})/a^+ = \|\mathfrak{C}(\tau)\|, \label{cummeqdist}\\
\Delta(\mathbf{P}_{a^-}, \mathbf{P}_{a^+}) &= \Delta(\mathbf{P}_{a^-}, \mathbf{P}_{a}) + \Delta(\mathbf{P}_{a}, \mathbf{P}_{a^+}). \label{cummcogeoln}
\end{align}
\end{subequations}
For~\eqref{cummeqdist}, the following can be shown
\begin{equation}
\langle \boldsymbol{\Psi}(\mathbf{P}_{a}), \mathbf{V}a^+\delta\boldsymbol{\Psi}_a \rangle_{\mathcal{H}} = \langle \boldsymbol{\Psi}(\mathbf{P}_{a}), \mathbf{V}a\delta\boldsymbol{\Psi}_{a^+} \rangle_{\mathcal{H}}.    
\end{equation}
Likewise, we can also obtain~\eqref{aricogeodist} for~\eqref{cummcogeoln}.

Consequently, it can be seen that the condition
$a^+\delta\boldsymbol{\Psi}_a = a\delta\boldsymbol{\Psi}_{a^+}$ is sufficient to make both \eqref{cummeqdist} and \eqref{cummcogeoln} hold.
By choosing $\delta\boldsymbol{\Psi}_a:=a\delta \boldsymbol{\Psi}_{\tau}$, the following cumulative pattern in the surrogate space then gives rise to~\eqref{geocumu}. 
\begin{equation}
\boldsymbol{\Psi}(\mathbf{P}_{a})  = \boldsymbol{\Psi}(\mathbf{P}_{a^-}) + a\delta \boldsymbol{\Psi}_{\tau}.
\end{equation}

\bibliographystyle{IEEEtran}
\bibliography{bibliography}

\begin{thebibliography}{10}
\providecommand{\url}[1]{#1}
\csname url@samestyle\endcsname
\providecommand{\newblock}{\relax}
\providecommand{\bibinfo}[2]{#2}
\providecommand{\BIBentrySTDinterwordspacing}{\spaceskip=0pt\relax}
\providecommand{\BIBentryALTinterwordstretchfactor}{4}
\providecommand{\BIBentryALTinterwordspacing}{\spaceskip=\fontdimen2\font plus
\BIBentryALTinterwordstretchfactor\fontdimen3\font minus \fontdimen4\font\relax}
\providecommand{\BIBforeignlanguage}[2]{{%
\expandafter\ifx\csname l@#1\endcsname\relax
\typeout{** WARNING: IEEEtran.bst: No hyphenation pattern has been}%
\typeout{** loaded for the language `#1'. Using the pattern for}%
\typeout{** the default language instead.}%
\else
\language=\csname l@#1\endcsname
\fi
#2}}
\providecommand{\BIBdecl}{\relax}
\BIBdecl

\bibitem{muddassir2021robotics}
M.~Muddassir, D.~Gómez~Domínguez, L.~Hu, S.~Chen, and D.~Navarro-Alarcon, ``Robotics meets cosmetic dermatology: Development of a novel vision-guided system for skin photo-rejuvenation,'' \emph{{IEEE/ASME} Trans. Mechatronics}, vol.~27, no.~2, pp. 666--677, 2022.

\bibitem{baumann2009cosmetic}
L.~S. Baumann and L.~Baumann, \emph{Cosmetic dermatology}.\hskip 1em plus 0.5em minus 0.4em\relax McGraw-Hill Professional Publishing, 2009.

\bibitem{9955367}
M.~Muddassir, G.~Limbert, B.~Zhang, A.~Duan, J.-J. Tan, and D.~Navarro-Alarcon, ``Model predictive thermal dose control of a robotic laser system to automate skin photorejuvenation,'' \emph{IEEE/ASME Transactions on Mechatronics}, vol.~28, no.~2, pp. 737--747, 2023.

\bibitem{ravichandar2020recent}
H.~Ravichandar, A.~S. Polydoros, S.~Chernova, and A.~Billard, ``Recent advances in robot learning from demonstration,'' \emph{Annual Review of Control, Robotics, and Autonomous Systems}, vol.~3, pp. 297--330, 2020.

\bibitem{10226460}
Z.~Jin, W.~Si, A.~Liu, W.-A. Zhang, L.~Yu, and C.~Yang, ``Learning a flexible neural energy function with a unique minimum for globally stable and accurate demonstration learning,'' \emph{IEEE Transactions on Robotics}, pp. 1--19, 2023.

\bibitem{Billard08chapter}
A.~Billard, S.~Calinon, R.~Dillmann, and S.~Schaal, ``Robot programming by demonstration,'' in \emph{Handbook of Robotics}, B.~Siciliano and O.~Khatib, Eds.\hskip 1em plus 0.5em minus 0.4em\relax Secaucus, NJ, USA: Springer, 2008, pp. 1371--1394.

\bibitem{10114055}
D.~Song, J.~Park, and Y.~J. Kim, ``{SSK}: Robotic pen-art system for large, nonplanar canvas,'' \emph{IEEE Transactions on Robotics}, vol.~39, no.~4, pp. 3106--3119, 2023.

\bibitem{liu2021robust}
R.~Liu, W.~Wan, K.~Koyama, and K.~Harada, ``Robust robotic 3-{D} drawing using closed-loop planning and online picked pens,'' \emph{IEEE Transactions on Robotics}, vol.~38, no.~3, pp. 1773--1792, 2021.

\bibitem{jafari2020surface}
B.~H. Jafari and N.~Gans, ``Surface parameterization and trajectory generation on regular surfaces with application in robot-guided deposition printing,'' \emph{IEEE Robotics and Automation Letters}, vol.~5, no.~4, pp. 6113--6120, 2020.

\bibitem{qureshi2021constrained}
A.~H. Qureshi, J.~Dong, A.~Baig, and M.~C. Yip, ``Constrained motion planning networks {X},'' \emph{IEEE Transactions on Robotics}, vol.~38, no.~2, pp. 868--886, 2021.

\bibitem{tiboni2023paintnet}
G.~Tiboni, R.~Camoriano, T.~Tommasi \emph{et~al.}, ``Paint{N}et: Unstructured multi-path learning from 3{D} point clouds for robotic spray painting,'' in \emph{IEEE/RSJ international conference on intelligent robots and systems}.\hskip 1em plus 0.5em minus 0.4em\relax IEEE, 2023.

\bibitem{liu2022robot}
P.~Liu, D.~Tateo, H.~B. Ammar, and J.~Peters, ``Robot reinforcement learning on the constraint manifold,'' in \emph{Conference on Robot Learning}.\hskip 1em plus 0.5em minus 0.4em\relax PMLR, 2022, pp. 1357--1366.

\bibitem{abu2020geometry}
F.~J. Abu-Dakka and V.~Kyrki, ``Geometry-aware dynamic movement primitives,'' in \emph{2020 IEEE International Conference on Robotics and Automation (ICRA)}.\hskip 1em plus 0.5em minus 0.4em\relax IEEE, 2020, pp. 4421--4426.

\bibitem{calinon2020gaussians}
S.~Calinon, ``Gaussians on {R}iemannian manifolds: Applications for robot learning and adaptive control,'' \emph{IEEE Robotics \& Automation Magazine}, vol.~27, no.~2, pp. 33--45, 2020.

\bibitem{rozo2022orientation}
L.~Rozo and V.~Dave, ``Orientation probabilistic movement primitives on riemannian manifolds,'' in \emph{Conference on Robot Learning}.\hskip 1em plus 0.5em minus 0.4em\relax PMLR, 2022, pp. 373--383.

\bibitem{doi:10.1177/02783649231204656}
\BIBentryALTinterwordspacing
A.~Duan, I.~Batzianoulis, R.~Camoriano, L.~Rosasco, D.~Pucci, and A.~Billard, ``A structured prediction approach for robot imitation learning,'' \emph{The International Journal of Robotics Research}, 2023. [Online]. Available: \url{https://doi.org/10.1177/02783649231204656}
\BIBentrySTDinterwordspacing

\bibitem{tenenbaum2000global}
J.~B. Tenenbaum, V.~d. Silva, and J.~C. Langford, ``A global geometric framework for nonlinear dimensionality reduction,'' \emph{science}, vol. 290, no. 5500, pp. 2319--2323, 2000.

\bibitem{shamai2018efficient}
G.~Shamai, M.~Zibulevsky, and R.~Kimmel, ``Efficient inter-geodesic distance computation and fast classical scaling,'' \emph{IEEE transactions on pattern analysis and machine intelligence}, 2018.

\bibitem{crane2013geodesics}
K.~Crane, C.~Weischedel, and M.~Wardetzky, ``Geodesics in heat: A new approach to computing distance based on heat flow,'' \emph{ACM Transactions on Graphics (TOG)}, vol.~32, no.~5, pp. 1--11, 2013.

\bibitem{jeong2008fast}
W.-K. Jeong and R.~T. Whitaker, ``A fast iterative method for {E}ikonal equations,'' \emph{SIAM Journal on Scientific Computing}, vol.~30, no.~5, pp. 2512--2534, 2008.

\bibitem{yang2022learning}
J.~Yang, J.~Zhang, C.~Settle, A.~Rai, R.~Antonova, and J.~Bohg, ``Learning periodic tasks from human demonstrations,'' in \emph{2022 International Conference on Robotics and Automation (ICRA)}.\hskip 1em plus 0.5em minus 0.4em\relax IEEE, 2022, pp. 8658--8665.

\bibitem{huang2020toward}
Y.~Huang, F.~J. Abu-Dakka, J.~Silv{\'e}rio, and D.~G. Caldwell, ``Toward orientation learning and adaptation in cartesian space,'' \emph{IEEE Transactions on Robotics}, vol.~37, no.~1, pp. 82--98, 2020.

\bibitem{li2020learning}
X.~Li, H.~Cheng, H.~Chen, and J.~Chen, ``Learning quasi-periodic robot motions from demonstration,'' \emph{Autonomous Robots}, vol.~44, pp. 251--266, 2020.

\bibitem{bakir2007predicting}
G.~BakIr, T.~Hofmann, A.~J. Smola, B.~Sch{\"o}lkopf, and B.~Taskar, \emph{Predicting structured data}.\hskip 1em plus 0.5em minus 0.4em\relax MIT press, 2007.

\bibitem{ciliberto2020general}
C.~Ciliberto, L.~Rosasco, and A.~Rudi, ``A general framework for consistent structured prediction with implicit loss embeddings.'' \emph{J. Mach. Learn. Res.}, vol.~21, no.~98, pp. 1--67, 2020.

\bibitem{alvarez2012kernels}
M.~Alvarez, L.~Rosasco, and N.~Lawrence, ``Kernels for vector-valued functions: A review,'' \emph{Foundations and Trends{\textregistered} in Machine Learning}, vol.~4, no.~3, pp. 195--266, 2012.

\bibitem{kimeldorf1970correspondence}
G.~S. Kimeldorf and G.~Wahba, ``A correspondence between {B}ayesian estimation on stochastic processes and smoothing by splines,'' \emph{The Annals of Mathematical Statistics}, vol.~41, no.~2, pp. 495--502, 1970.

\bibitem{duvenaud2014automatic}
D.~Duvenaud, ``Automatic model construction with {G}aussian processes,'' Ph.D. dissertation, University of Cambridge, 2014.

\bibitem{mackay1998introduction}
D.~J. MacKay, ``Introduction to {G}aussian processes,'' \emph{{NATO ASI} series {F}: {C}omputer and systems sciences}, vol. 168, pp. 133--166, 1998.

\bibitem{noorzadeh2014modeling}
S.~Noorzadeh, M.~Niknazar, B.~Rivet, J.~Fontecave-Jallon, P.-Y. Gum{\'e}ry, and C.~Jutten, ``Modeling quasi-periodic signals by a non-parametric model: Application on fetal {ECG} extraction,'' in \emph{2014 36th Annual International Conference of the IEEE Engineering in Medicine and Biology Society}.\hskip 1em plus 0.5em minus 0.4em\relax IEEE, 2014, pp. 1889--1892.

\bibitem{angus2018inferring}
R.~Angus, T.~Morton, S.~Aigrain, D.~Foreman-Mackey, and V.~Rajpaul, ``Inferring probabilistic stellar rotation periods using {G}aussian processes,'' \emph{Monthly Notices of the Royal Astronomical Society}, vol. 474, no.~2, pp. 2094--2108, 2018.

\bibitem{bose2011survey}
P.~Bose, A.~Maheshwari, C.~Shu, and S.~Wuhrer, ``A survey of geodesic paths on 3{D} surfaces,'' \emph{Computational Geometry}, vol.~44, no.~9, pp. 486--498, 2011.

\bibitem{10.1111/rssb.12508}
\BIBentryALTinterwordspacing
D.~Li, M.~Mukhopadhyay, and D.~B. Dunson, ``{Efficient manifold approximation with spherelets},'' \emph{Journal of the Royal Statistical Society Series B: Statistical Methodology}, vol.~84, no.~4, pp. 1129--1149, 04 2022. [Online]. Available: \url{https://doi.org/10.1111/rssb.12508}
\BIBentrySTDinterwordspacing

\bibitem{absil2008optimization}
P.-A. Absil, R.~Mahony, and R.~Sepulchre, \emph{Optimization algorithms on matrix manifolds}.\hskip 1em plus 0.5em minus 0.4em\relax Princeton University Press, 2008.

\bibitem{wahba1990spline}
G.~Wahba, \emph{Spline models for observational data}.\hskip 1em plus 0.5em minus 0.4em\relax Siam, 1990, vol.~59.

\bibitem{schulman2016learning}
J.~Schulman, J.~Ho, C.~Lee, and P.~Abbeel, ``Learning from demonstrations through the use of non-rigid registration,'' in \emph{Robotics Research}.\hskip 1em plus 0.5em minus 0.4em\relax Springer, 2016, pp. 339--354.

\bibitem{cai2021landmark}
H.~Cai, Y.~Guo, Z.~Peng, and J.~Zhang, ``Landmark detection and 3{D} face reconstruction for caricature using a nonlinear parametric model,'' \emph{Graphical Models}, vol. 115, p. 101103, 2021.

\bibitem{zhou2023neural}
P.~Zhou, P.~Zheng, J.~Qi, C.~Li, A.~Duan, M.~Xu, V.~Wu, and D.~Navarro-Alarcon, ``Neural reactive path planning with {R}iemannian motion policies for robotic silicone sealing,'' \emph{Robotics and Computer-Integrated Manufacturing}, vol.~81, p. 102518, 2023.

\bibitem{luyin_tro2022}
L.~Hu, D.~Navarro-Alarcon, A.~Cherubini, M.~Li, and L.~Li, ``On radiation-based thermal servoing: New models, controls, and experiments,'' \emph{{IEEE} Transactions on Robotics}, vol.~38, no.~3, pp. 1945--1958, 2022.

\end{thebibliography}
	
\vfill
\end{document}